\documentclass[letterpaper, 10 pt, conference]{ieeeconf}
\IEEEoverridecommandlockouts    %
\usepackage{hyperref}
\usepackage{threeparttable}
\usepackage{algorithm}
\usepackage{cite}

\hypersetup{
    colorlinks=true,
    linkcolor=magenta,
    filecolor=magenta,      
    urlcolor=magenta,
}

\newcommand{\mathbfz}{\mathbf{z}}

\usepackage{wrapfig}
\usepackage{lipsum}
\usepackage{xspace}
\newcommand{\etal}{\emph{et al.}\xspace}
\usepackage{multirow}
\usepackage[table]{xcolor} %
\definecolor{lightgray}{RGB}{220, 220, 220} %
\usepackage{caption}
\usepackage{subcaption}
\newcommand{\smallR}{{\scriptstyle{\text{R}}}}
\newcommand{\smallL}{{\scriptstyle{\text{L}}}}

\newcommand{\tinytext}[1]{\scalebox{0.8}{$#1$}}
\newcommand{\tinytinytext}[1]{\scalebox{0.75}{$#1$}}

\usepackage{graphics}           
\usepackage{times}              
\usepackage{amsmath}            
\usepackage{amssymb}            
\usepackage{graphicx}
\usepackage{algorithm}
\usepackage[noend]{algpseudocode}
\usepackage{booktabs}
\usepackage{color}
\definecolor{instructioncolor}{rgb}{.5,.5,.5}

\usepackage[font=small]{caption}

\def\eqref#1{Eq.~(\ref{#1})}

\makeatletter
\usepackage{xspace}
\DeclareRobustCommand\onedot{\futurelet\@let@token\@onedot}
\def\@onedot{\ifx\@let@token.\else.\null\fi\xspace}

\def\etal{{et al}\onedot}
\makeatother

\usepackage{array}
\newcolumntype{L}[1]{>{\raggedright\let\newline\\\arraybackslash\hspace{0pt}}m{#1}}
\newcolumntype{C}[1]{>{\centering\let\newline\\\arraybackslash\hspace{0pt}}m{#1}}
\newcolumntype{R}[1]{>{\raggedleft\let\newline\\\arraybackslash\hspace{0pt}}m{#1}}

\usepackage{bm}
\usepackage{multirow}
\usepackage{rotating}  %

\renewcommand{\thefigure}{\Roman{figure}}

\title{\LARGE \bf Diff-IP2D: Diffusion-Based Hand-Object Interaction Prediction \\ on Egocentric Videos}

\author{Junyi~Ma$^1$, Xieyuanli~Chen$^2$, Jingyi~Xu$^3$, Hesheng~Wang$^{1*}$
\thanks{$^{1}$Junyi~Ma and Hesheng~Wang are with IRMV Lab, the Department of Automation, Shanghai Jiao Tong University, Shanghai 200240, China.}
\thanks{$^{2}$Xieyuanli~Chen is with the College of Intelligence Science and Technology, National University of Defense Technology, Changsha 410073, China.}
\thanks{$^3$Jingyi~Xu is with the Department of Electronic Engineering, Shanghai Jiao Tong University, Shanghai 200240, China. }
\thanks{$^{*}$Corresponding author email: wanghesheng@sjtu.edu.cn}
}

\begin{document}
\maketitle

\IEEEpeerreviewmaketitle
\thispagestyle{empty}
\pagestyle{empty}

\begin{abstract}
Understanding how humans would behave during hand-object interaction (HOI) is vital for applications in service robot manipulation and extended reality. 
To achieve this, some recent works simultaneously forecast hand trajectories and object affordances on human egocentric videos. The joint prediction serves as a comprehensive representation of future HOI in 2D space, indicating potential human motion and motivation.
However, the existing approaches mostly adopt the autoregressive paradigm, which lacks bidirectional constraints within the holistic future sequence, and accumulates errors along the time axis.
Meanwhile, they overlook the effect of camera egomotion on first-person view predictions. To address these limitations, we propose a novel diffusion-based HOI prediction method, namely Diff-IP2D, to forecast future hand trajectories and object affordances with bidirectional constraints in an iterative non-autoregressive manner on egocentric videos.
Motion features are further integrated into the conditional denoising process to enable Diff-IP2D aware of the camera wearer's dynamics for more accurate interaction prediction.
Extensive experiments demonstrate that Diff-IP2D significantly outperforms the state-of-the-art baselines on both the off-the-shelf and our newly proposed evaluation metrics.
This highlights the efficacy of leveraging a generative paradigm for 2D HOI prediction. The code of Diff-IP2D is released as open source at \url{https://github.com/IRMVLab/Diff-IP2D}.
\end{abstract}

\section{Introduction}
\label{sec:intro}
Accurately anticipating human intentions and future actions is important for artificial intelligence systems in robotics and extended reality \cite{bahl2023affordances,chang2023look,han2022umetrack}. Recent works have tried to tackle the problem from various perspectives, including action recognition and anticipation \cite{xu2023dynamic,zhang2024object,zheng2023icme}, gaze prediction \cite{zhang2017deep,lai2024eye,li2018eye}, hand trajectory prediction \cite{liu2022joint,bao2023uncertainty,liu2020forecasting,shan2020understanding}, and object affordance extraction \cite{liu2022joint,ye2023affordance,liu2020forecasting,xu2023interdiff}. 
Among them, jointly predicting hand motion and object affordances can effectively facilitate more reasonable robot manipulation as the prior contextual information, which has been demonstrated on some robot platforms \cite{bahl2023affordances,mendonca2023structured,bahl2022human}. We believe that deploying such models pretrained by internet-scale human videos on robots is a promising path towards embodied agents.
Therefore, our work aims to jointly predict hand trajectories and object affordances on egocentric videos as a concrete hand-object interaction (HOI) expression, following the problem modeling of the previous works~\cite{liu2022joint, liu2020forecasting}.

\begin{figure}
  \centering
  \includegraphics[width=1\linewidth]{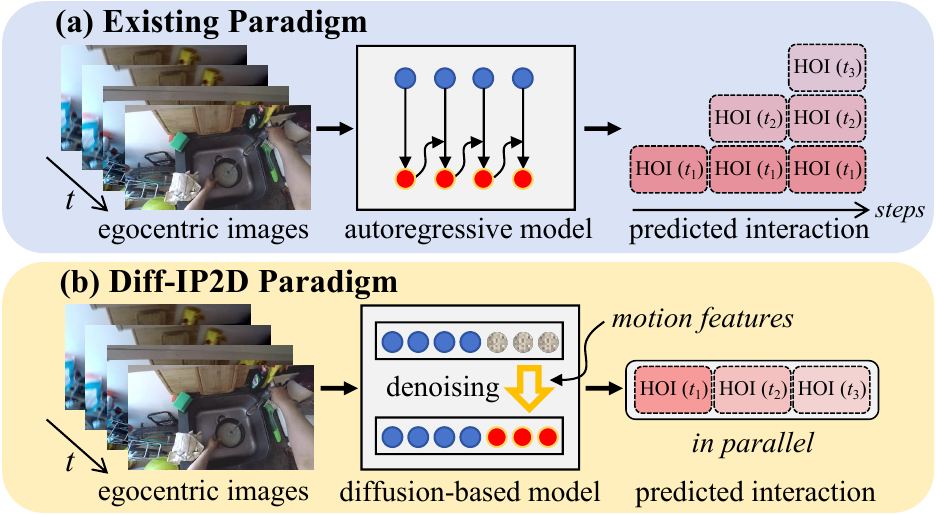}
  \caption{Diff-IP2D~vs.~Existing Paradigm. The existing HOI prediction paradigm (a) tends to accumulate prediction errors under unidirectional constraints. In contrast, our proposed Diff-IP2D (b) directly forecasts all future HOI states in parallel with motion-aware denoising diffusion, mitigating error accumulation with bidirectional constraints.}
  \label{fig:motivation_maintext}
  \vspace{-0.7cm}
\end{figure}

Currently, the state-of-the-art (SOTA) approaches \cite{liu2022joint,bao2023uncertainty} predicting hand trajectories and object affordances on egocentric videos tend to exploit the autoregressive (AR) model. 
They reason about the next HOI state only according to the previous steps (Fig.~\ref{fig:motivation_maintext}(a)), focusing on unidirectionally forward constraints with \textit{temporal causality}. 
However, expected ``post-contact states'' also affect ``pre-contact states'' following \textit{spatial causality} with human intentions that persist across the holistic HOI process as an oracle (also refer to the cup example in Sec.~2 of the supplementary material). 
Inspired by this, we argue that predicting future HOI states in parallel attending to bidirectional constraints within the holistic sequence outperforms generating the next state autoregressively and mitigates temporal error accumulation.
With diffusion models emerging across multiple domains \cite{ho2020denoising,dhariwal2021diffusion,esser2023structure,ji2023ddp,liu2023difflow3d,peebles2023scalable,gong2022diffuseq,gong2023diffuseq}, 
their strong forecasting capability has been widely validated. 
Therefore, we propose a diffusion-based method to predict future hand-object interaction in an iterative non-autoregressive (iter-NAR) manner (Fig.~\ref{fig:motivation_maintext}(b)), considering bidirectional constraints in the latent space compared to traditional AR generation.

Moreover, we also find two inherent gaps affecting HOI prediction in the existing paradigm: 1) Directly predicting the projection of 3D future hand trajectories and object affordances on 2D egocentric image plane is an ill-posed problem involving spatial ambiguities. There is generally a gap between 2D pixel movements and 3D real actions, which can be bridged by spatial transformation across multiple views changing with egomotion. 2) The past egocentric videos are absorbed to predict future interaction states on the last observed image, which is actually a ``canvas'' from a different view w.r.t all the other frames. Therefore, there is also a gap between the last observation (egocentric view) and the other observations (analogous to exocentric view) caused by egomotion.
To fill these two gaps, we further propose integrating the camera wearer's egomotion into our diffusion-based paradigm by the cross-attention between egomotion homography features and HOI features. It enables the denoising model aware of the camera wearer's dynamics and the spatial relationship between consecutive frames.

The main contributions of this paper are as follows: 1) We propose a \underline{diff}usion-based hand-object \underline{i}nteraction \underline{p}rediction method, dubbed Diff-IP2D. To our best knowledge, this is the first work using the devised denoising diffusion probabilistic model to jointly forecast future hand trajectories and object affordances with only \underline{2D} egocentric videos as input. It provides a foundation iter-NAR generative paradigm in the field of HOI prediction. 2) The homography egomotion features are integrated to fill the motion-related gaps inherent in HOI prediction on egocentric videos. 3) Comprehensive experiments conducted on our extended evaluation metrics demonstrate that Diff-IP2D can predict better hand trajectories and object affordances compared to the SOTA baselines, showing its potential for deployment on artificial intelligence systems.

\vspace{-0.1cm}
\section{Related work}
\label{sec:related_work}

\textbf{Understanding hand-object interaction.} Human HOI comprehension can guide the downstream tasks in artificial intelligence systems. Calway \etal \cite{calway2015discovering} and Liu \etal \cite{Liu_2017_ICCV} both innovatively connect human tasks to relevant objects, which underlines the relationship between object-centric interaction and goal-oriented human activities. After that, many works contribute to HOI understanding by pixel-wise segmentation \cite{schroder2017hand,darkhalil2022epic,zhang2022fine,Higgins_2023_CVPR}, bounding-box-wise detection \cite{furnari2020rolling,fan2021understanding,Shiota_2024_WACV,shan2020understanding}, fine-grained hand/object pose estimation \cite{romero2022embodied,zhou2020monocular,Lin_2023_CVPR,yang2022artiboost,liu2021semi,lim2013parsing}. Ego4D \cite{grauman2022ego4d} further provides a standard benchmark that divides HOI understanding into several predefined subtasks.

\textbf{Predicting hand-object interaction.} Analyzing only past human behavior may be insufficient for service robot manipulation or extended reality. Forecasting possible future object-centric HOI states based on historical observations is also valuable, which attracts increasing attention due to the general knowledge that can be transferred to robot applications \cite{bahl2023affordances,mendonca2023structured,bahl2022human,bharadhwaj2023towards}. 
For example, Dessalene \etal \cite{Dessalene2023Forecasting} propose to generate contact anticipation maps and next active object segmentations as future HOI predictions. Liu \etal \cite{liu2020forecasting} first achieve hand trajectory and object affordance prediction simultaneously, revealing that predicting hand motion benefits the extraction of interaction hotspots, and vice versa. Following this work, Liu \etal \cite{liu2022joint} further develop an object-centric Transformer to jointly forecast future trajectories and affordances autoregressively, and annotate publicly available datasets to support future works. More recently, Bao \etal \cite{bao2023uncertainty} lift the problem to 3D spaces where hand trajectories are predicted by an uncertainty-aware state space Transformer in an autoregressive manner. However, it needs additional 3D perception inputs from the RGB-D camera.
In this work, we still achieve joint hand trajectory and object affordance prediction on 2D human videos rather than in 3D space. We focus on capturing more general knowledge from only egocentric 2D observations in an iterative non-autoregressive manner, rather than the autoregressive way of the SOTA works \cite{liu2022joint,bao2023uncertainty}. 

\textbf{Diffusion-based egocentric video analysis.} Diffusion models have been successfully utilized in some egocentric vision tasks due to their strong generation ability, such as video prediction \cite{luo2024put,chang2023look}, human mesh recovery \cite{zhang2023probabilistic,liu2023egohmr}, 3D HOI reconstruction \cite{Ye_2023_ICCV,zhu2023get}, and 3D HOI synthesizing \cite{ye2023affordance,zhang2024hoidiffusion}. However, none of these works concentrate on the combination of fine-grained hand trajectories and object affordances as future HOI representations for potential utilization in artificial intelligence systems. Our Diff-IP2D first achieves this based on the denoising diffusion probabilistic model \cite{ho2020denoising}, which dominates the existing paradigm \cite{liu2022joint,bao2023uncertainty} in prediction performance on egocentric videos.

\begin{figure*}
  \centering
  \includegraphics[width=1\linewidth]{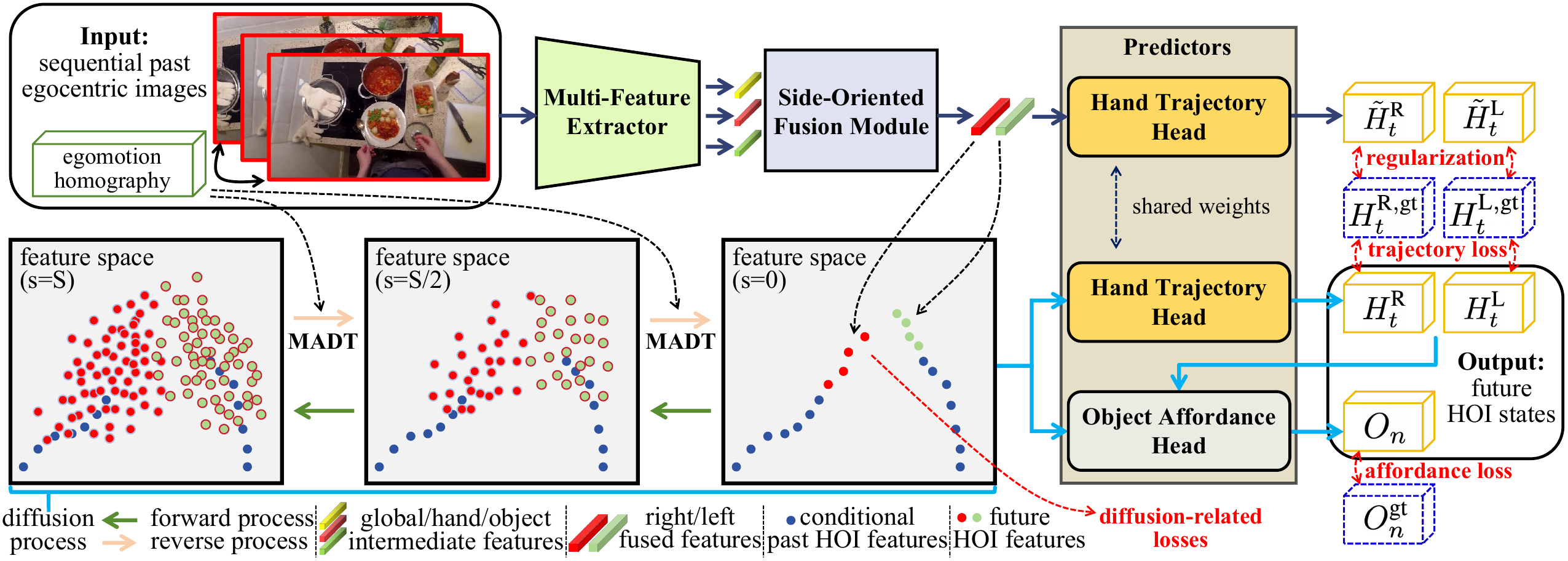}
  \caption{System overview of Diff-IP2D. Our proposed paradigm takes in sequential past egocentric images and jointly predicts hand trajectories and object affordances as future HOI states. The observations are mapped to the latent feature space for the diffusion process.}
  \label{fig:sytem_overview}
  \vspace{-0.5cm}
\end{figure*}

\section{Proposed Method}
\label{sec:proposed_method}

In this section, we first introduce the preliminaries in Sec.~\ref{sec:preliminary}. Then we elaborate on our proposed Diff-IP2D architecture in Sec.~\ref{sec:architecture}. Next, we clarify the training and inference schemes in Sec.~\ref{sec:training} and Sec.~\ref{sec:inference} respectively.

\subsection{Preliminaries}
\label{sec:preliminary}
\textbf{Task definition.} Given the video clip of past egocentric observations $\mathcal{I}=\{I_t\}_{t=-N_\text{p}+1}^{0}$, we aim to predict future hand trajectories $\mathcal{H}=\{H_t^\smallR,H_t^\smallL\}_{t=1}^{N_\text{f}} (H_t^\smallR,H_t^\smallL \in \mathbb{R}^{2})$ and potential object contact points $\mathcal{O}=\{O_n\}_{n=1}^{N_\text{o}} (O_n \in \mathbb{R}^{2})$, where $N_\text{p}$ and $N_\text{f}$ are the numbers of frames in the past and future time horizons respectively, and $N_\text{o}$ denotes the number of predicted contact points used to calculate interaction hotspots as object affordances. Following the previous works \cite{liu2022joint,liu2020forecasting}, we predict future positions of the right hand, left hand, and affordance of the next active object on the last observed image of the input videos as a canvas.

\textbf{Diffusion models.} In this work, we propose a diffusion-based approach to gradually corrupt the input to noisy features and then train a denoising model to reverse this process. We first map the input images into a latent space $\mathbfz_0\sim q(\mathbfz_0)$, which is then corrupted to a standard Gaussian noise $\mathbfz_S \sim \mathcal{N}(0,\textbf{\text{I}})$. In the forward process, the perturbation operation can be represented as $q(\mathbfz_s|\mathbfz_{s-1}) = \mathcal{N}(\mathbfz_s;\sqrt{1-\beta_s}\mathbfz_{s-1},\beta_s\textbf{\text{I}})$, where $\beta_s$ is the predefined variance scale. In the reverse process, we set a denoising diffusion model to gradually reconstruct the latent $\mathbfz_0$ from the noisy $\mathbfz_S$. The denoised features can be used to recover the final future hand trajectories and object affordances.

\subsection{Architecture}
\label{sec:architecture}
\textbf{System overview.} Accurately reconstructing the future part of the input sequence is critical in diffusion-based prediction. We found that ground-truth hand waypoints $\mathcal{H}^\text{gt}=\{H_t^{\smallR,\text{gt}},H_t^{\smallL,\text{gt}}\}_{t=1}^{N_\text{f}} (H_t^{\smallR,\text{gt}},H_t^{\smallL,\text{gt}} \in \mathbb{R}^{2})$ and contact points $\mathcal{O}^{\text{gt}}=\{O^{\text{gt}}_n\}_{n=1}^{N_\text{o}} (O^{\text{gt}}_n \in \mathbb{R}^{2})$ provide discrete and sparse supervision for reconstruction, which is not enough for capturing high-level semantics such as human intentions in the denoising process. Therefore, as Fig.~\ref{fig:sytem_overview} shows, we first use the Multi-Feature Extractor (MFE) and Side-Oriented Fusion Module (SOFM) to transform input images into latent HOI features, and then implement diffusion-related operation in the latent continuous space. The HOI features denoised by the Motion-Aware Denoising Transformer (MADT) are further absorbed by the Hand Trajectory Head and Object Affordance Head to generate future hand trajectories and object hotspots.

\textbf{Multi-Feature Extractor (MFE).} Following the previous work \cite{liu2022joint}, we use MFE that consists of a pretrained Temporal Segment Network (TSN) provided by Furnari \etal \cite{furnari2020rolling}, RoIAlign \cite{he2017mask} with average pooling, and Multilayer Perceptron (MLP) to extract hand, object, and global 
intermediate features for each sequence image $I_t \in \mathcal{I}$. The positions of hand-object bounding boxes detected by the off-the-shelf approach \cite{shan2020understanding} are also encoded to feature vectors fused with hand and object intermediate features. That is, all the following HOI features in this work encompass spatial information of hands and objects within each image. 

\textbf{Side-Oriented Fusion Module (SOFM).} Our proposed SOFM is a learnable linear transformation to fuse the above-mentioned three types of feature vectors into the final latent form for two sides respectively. Specifically, the global features and right-side features (right-hand/object features) are concatenated and are then linearly transformed to the right-side HOI features $\mathcal{F}^\smallR=\{F_t^\smallR\}_{t=-N_\text{p}+1}^{X} (F_t^\smallR \in \mathbb{R}^{a}, X=N_\text{f}$ for training and $X=0$ for inference). The operation and feature sizes are the same as the left-side counterparts, leading to \smash{$\mathcal{F}^\smallL=\{F_t^\smallL\}_{t=-N_\text{p}+1}^{X}$}. We further concatenate the side-oriented features along the time axis respectively to generate the input latents $F^\smallR_{\text{seq}}$, $F^\smallL_{\text{seq}} \in \mathbb{R}^{(N_\text{p}+X)\times a}$ for the following diffusion model.

\begin{figure}
  \centering
\includegraphics[width=0.9\linewidth]{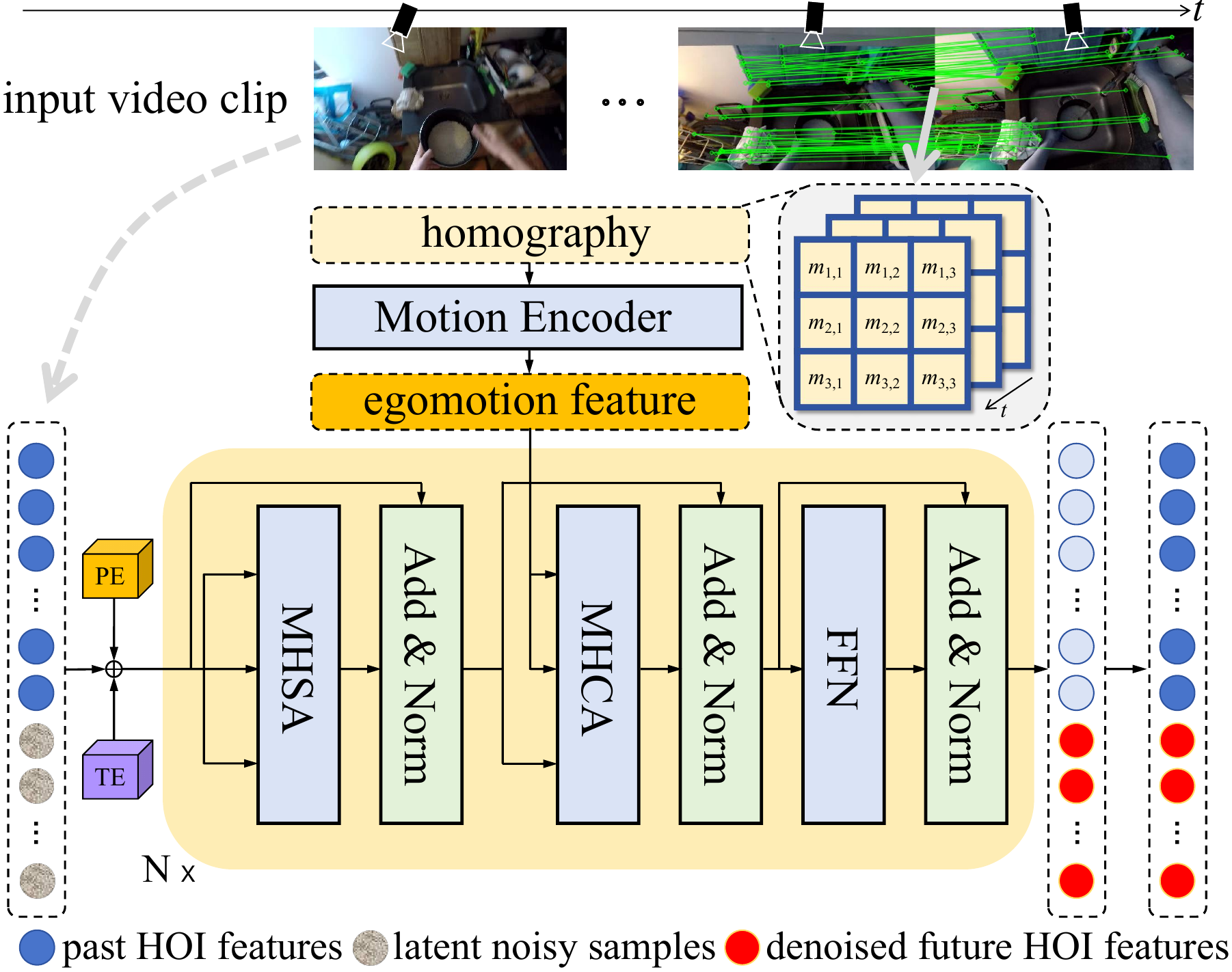}
\vspace{-0.2cm}
  \caption{MADT architecture. It receives corrupted HOI latents with position embedding (PE) and time embedding (TE), and outputs denoised future HOI latents under egomotion guidance.}
  \label{fig:transformer}
  \vspace{-0.6cm}
\end{figure}

\textbf{Motion-Aware Denoising Transformer (MADT).} Our proposed MADT takes in the noisy latent HOI features and reconstructs future HOI features for the following predictors conditioned on past HOI counterparts. MADT consists of devised Transformer layers as shown in Fig.~\ref{fig:transformer}, thus imposing bidirectional constraints on temporal latents. Following the previous work \cite{gong2022diffuseq}, we anchor the past HOI features for both forward and reverse processes. We only impose noises and denoise at the positions of the future feature sequence. The features of the two sides are denoised using the same model, leading to \smash{$\hat{F}^\smallR_{\text{seq}}$} and \smash{$\hat{F}^\smallL_{\text{seq}}$}.
In addition, egomotion guidance is proposed here to fill the gaps mentioned in Sec.~\ref{sec:intro}.
Specifically, we first extract SIFT~\cite{lowe2004distinctive} descriptors to find the pixel correspondence between two adjacent past images in $\mathcal{I}$. Then we use RANSAC~\cite{fischler1981random} to solve the homography matrix that maximizes the number of inliers in keypoint pairs. We accumulate the consecutive homography matrices and obtain 
$M_{\text{seq}} \in \mathbb{R}^{N_\text{p}\times 3\times 3}$
representing the camera wearer's motion between $I_t$ ($t\leq0$) and $I_0$.
They are further linearly embedded into an egomotion feature $E_{\text{seq}} \in \mathbb{R}^{N_\text{p}\times b}$ by Motion Encoder. The multi-head cross-attention module (MHCA) in MADT then absorbs the egomotion features to guide the denoising process. More analysis on the use of egomotion guidance can be found in Sec.~1 of the supplementary material.

\textbf{Predictors.} Our proposed predictors consist of Hand Trajectory Head (HTH) and Object Affordance Head (OAH). HTH contains an MLP that receives the future parts of the denoised features $\hat{F}^\smallR_{\text{seq}}[N_\text{p}\!+\!1\!:\!N_\text{p}\!+\!N_\text{f}]$ and $\hat{F}^\smallL_{\text{seq}}[N_\text{p}\!+\!1\!:\!N_\text{p}\!+\!N_\text{f}]$, to generate future hand waypoints $\mathcal{H}$ of two hands. As to OAH, we empirically exploit Conditional Variational Autoencoder (C-VAE) \cite{sohn2015learning} to generate possible contact points $\mathcal{O}$ of the next active object in near future. Taking the right hand as an example, the condition is selected as the time-averaged \smash{$\hat{F}^\smallR_{\text{seq}}$} and predicted waypoints \smash{$H_t^\smallR$}. Note that we additionally consider denoised future HOI features $\hat{F}^\smallR_{\text{seq}}[N_\text{p}\!+\!1:N_\text{p}\!+\!N_\text{f}]$ ($t\!>\!0$) besides the features from past observations ($t\!\leq\!0$) for object affordance prediction. This aligns with the intuitive relationship between the contact points and the overall interaction process.

\subsection{Training}
\label{sec:training}
\textbf{Forward process.} We implement partial noising \cite{gong2022diffuseq} in the forward process during training. Taking the right side as an example, the output of SOFM is first extended by a Markov transition $q(\mathbfz_0|F^\smallR_{\text{seq}})= \mathcal{N}(F^\smallR_{\text{seq}},\beta_0\textbf{\text{I}})$, where $F^\smallR_{\text{seq}} \in \mathbb{R}^{(N_\text{p}+N_\text{f})\times a}$. In each following forward diffusion step, we implement $q(\mathbfz_s|\mathbfz_{s-1})$ by adding noise to the future part of $\mathbfz_{s-1}$, i.e., $\mathbfz_{s-1}[N_\text{p}\!+\!1\!:\!N_\text{p}\!+\!N_\text{f}]$ for both sides. 

\textbf{Reverse process.} After corrupting $\mathbfz_0$ to $\mathbfz_S$ by the forward process, our proposed MADT is adopted to denoise $\mathbfz_S$ to $\mathbfz_0$. Considering the proposed guidance of egomotion features, the reverse process can be modeled as $p_{\scriptscriptstyle{\text{MADT}}}(\mathbfz_{0:S}):=p(\mathbfz_S)\prod_{s=1}^{S}p_{\scriptscriptstyle{\text{MADT}}}(\mathbfz_{s-1}|\mathbfz_{s},M_{\text{seq}})$. Specifically, the MADT model $f_{\scriptscriptstyle\text{MADT}}(\mathbfz_s,s,M_{\text{seq}})$ predicts the injected noise for each forward step with $p_{\scriptscriptstyle{\text{MADT}}}(\mathbfz_{s-1}|\mathbfz_{s},M_{\text{seq}})=\mathcal{N}(\mathbfz_{s-1};\mu_{\scriptscriptstyle{\text{MADT}}}(\mathbfz_s,s,M_{\text{seq}}),\sigma_{\scriptscriptstyle{\text{MADT}}}(\mathbfz_s,s,M_{\text{seq}}))$. 
The same denoising operation and motion-aware guidance are applied to HOI features of both sides.

\textbf{Training objectives.} The loss function training Diff-IP2D comprises four components: diffusion-related losses, trajectory loss, affordance loss, and a regularization term (see Fig.~\ref{fig:sytem_overview}). Taking the right side as an example, we use the variational lower bound $\mathcal{L}_\text{VLB}^\smallR$ as diffusion-related losses:
\begin{align}
\textstyle
\mathcal{L}_\text{VLB}^\smallR = \sum_{s=2}^S||\mathbfz_0^\smallR-f_{\scriptscriptstyle\text{MADT}}(\mathbfz_s^\smallR,s,M_{\text{seq}})||^2 \nonumber\\ + ||F_{\text{seq}}^\smallR-\hat{F}_{\text{seq}}^\smallR||^2,
\label{eq:vlb_loss}
\end{align}
where $\hat{F}_{\text{seq}}^\smallR = f_{\scriptscriptstyle\text{MADT}}(\mathbfz_1^\smallR,1,M_{\text{seq}})$. To reconstruct hand trajectories beyond the latent feature space, we further set trajectory loss $\mathcal{L}_\text{traj}^\smallR$ with the distance between the ground-truth waypoints and the ones predicted by HTH:
\begin{align}    
    \textstyle
    \mathcal{L}_\text{traj}^\smallR = \sum_{t=1}^{N_\text{f}}||H_t^\smallR-H_t^{\smallR,\text{gt}}||^2,
\label{eq:hand_loss}
\end{align}
where $H_t^\smallR=f_{\scriptscriptstyle\text{HTH}}(\hat{F}_{\text{seq}}^\smallR[N_\text{p}\!+\!1\!:\!N_\text{p}\!+\!N_\text{f}])$. 
As to the object affordance prediction, we also compute the affordance loss $\mathcal{L}_\text{aff}$ after multiple stochastic sampling considering the next active object recognized following Liu \etal \cite{liu2022joint} (assuming in the right side here for brevity):
\begin{align}
    \textstyle
    \mathcal{L}_\text{aff} = \sum_{n=1}^{N_\text{o}}||O_n-O_n^{\text{gt}}||^2 + c\mathcal{L}_\text{KL},
\label{eq:obj_loss}
\end{align}
where $O_n\!=\!f_{\scriptscriptstyle\text{OAH}}(\hat{F}_{\text{seq}}^\smallR,H_t^\smallR)$ is the predicted contact points, and $\mathcal{L}_\text{KL}=\frac{1}{2}(-\log \sigma^2_{\scriptscriptstyle\text{OAH}}(\hat{F}_{\text{seq}}^\smallR,H_t^\smallR)+\mu^2_{\scriptscriptstyle\text{OAH}}(\hat{F}_{\text{seq}}^\smallR,H_t^\smallR)+\sigma^2_{\scriptscriptstyle\text{OAH}}(\hat{F}_{\text{seq}}^\smallR,H_t^\smallR)-1)$ is the KL-Divergence regularization for C-VAE, which is scaled by $c=1e\text{-}3$. The latent features and predicted hand waypoints are fused by MLP suggested by the previous work \cite{liu2022joint}. We consider both reconstructed future HOI features $\hat{F}^\smallR_{\text{seq}}[N_\text{p}\!+\!1\!:\!N_\text{p}\!+\!N_\text{f}]$ and anchored past counterparts $\hat{F}^\smallR_{\text{seq}}[0\!:\!N_\text{p}]$ compared to \cite{liu2022joint} as mentioned before.
We also notice that the latent feature spaces before and after the denoising diffusion process represent the same ``profile'' of the input HOI sequence. Therefore, we propose an additional regularization term implicitly linking $F^\smallR_{\text{seq}}$ and $\hat{F}^\smallR_{\text{seq}}$ by hand trajectory prediction:
\begin{align}
    \textstyle
    \mathcal{L}_\text{reg}^\smallR = \sum_{t=1}^{N_\text{f}}||\tilde{H}_t^\smallR-H_t^{\smallR,\text{gt}}||^2,
\label{eq:reg_loss}
\end{align}
where $\tilde{H}_t^\smallR=f_{\scriptscriptstyle\text{HTH}}(F_{\text{seq}}^\smallR[N_\text{p}\!+\!1\!:\!N_\text{p}\!+\!N_\text{f}])$. Although Eq.~(\ref{eq:reg_loss}) does not explicitly contain the term $\hat{F}^\smallR_{\text{seq}}$, the training direction is the same with Eq.~(\ref{eq:hand_loss}), thus maintaining training stability.
The regularization helps distill HOI state knowledge by building a closer gradient connection constraining the two latent spaces alongside the diffusion process for better optimization. Here we do not use object affordance prediction for regularization because we empirically found that incorporating OAH mitigates training efficiency while the positive effect is not obvious. Sec.~3 in the supplementary material provides more detailed clarification about the motivation of our proposed regularization strategy.
Finally, we get the total loss $\mathcal{L}_\text{total}$, the weighted sum of all the above-mentioned losses to train our proposed Diff-IP2D. Besides, we leverage the importance sampling technique proposed in improved DDPM \cite{nichol2021improved}, which promotes the training process focusing more on the steps with relatively large $\mathcal{L}_\text{total}$.

\begin{figure}[t]
  \centering
  \includegraphics[width=1\linewidth]{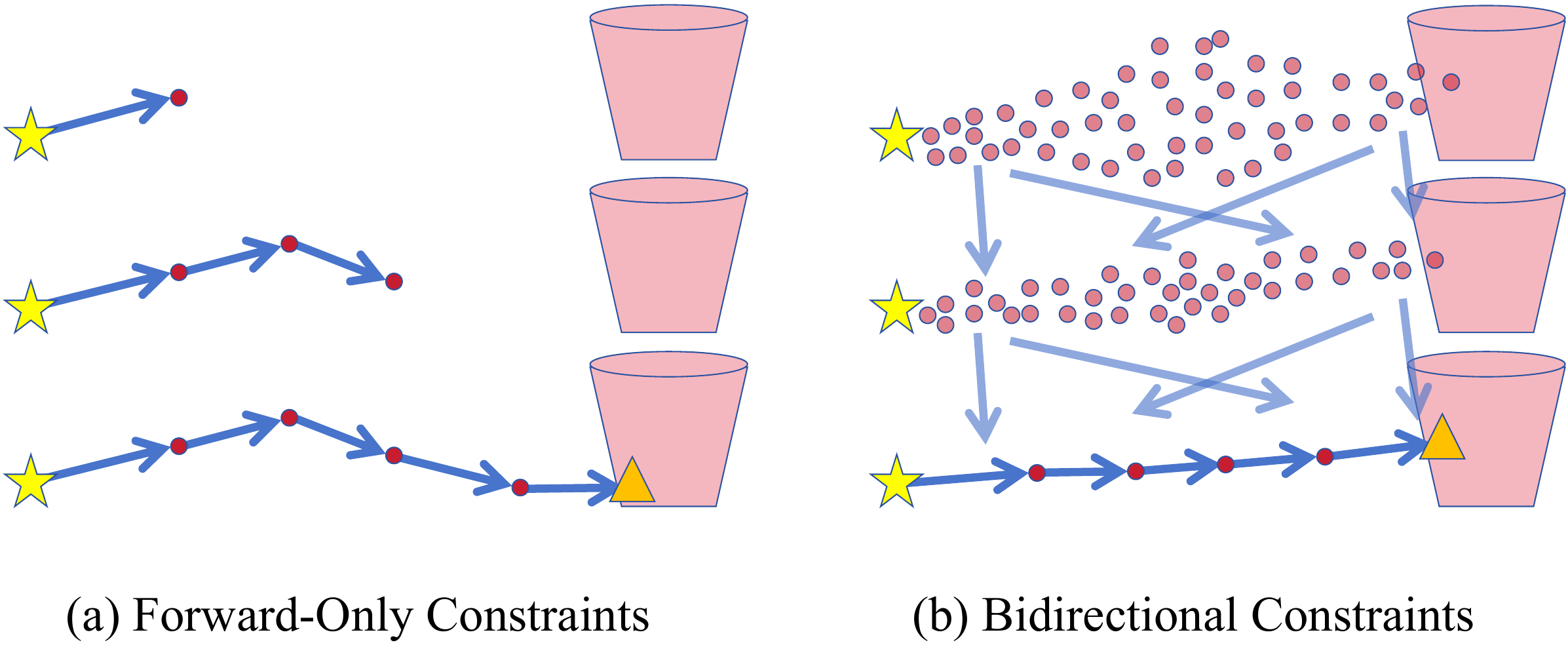}
  \caption{Comparison of AR prediction with forward-only constraints and our iter-NAR prediction with both forward and backward constraints. The yellow star represents the starting point of trajectory prediction, and the orange triangle represents the interaction point with the target object.}
  \label{fig:compare_ar}
  \vspace{-0.7cm}
\end{figure}

\vspace{-0.1cm}
\subsection{Inference}
\label{sec:inference}
\textbf{Prediction pipeline.} In the inference stage, we first sample $F^\smallR_{\text{noise}}$, $F^\smallL_{\text{noise}} \in \mathbb{R}^{N_\text{f}\times a}$ from a standard Gaussian distribution, which are then concatenated with $F^\smallR_{\text{seq}}$, $F^\smallL_{\text{seq}} \in \mathbb{R}^{N_\text{p}\times a}$ along the time axis to generate $\mathbfz_S^\smallR$ and $\mathbfz_S^\smallL$. Then we use MADT to predict $\mathbfz_0^\smallR$ and $\mathbfz_0^\smallL$ based on DDIM sampling \cite{song2020denoising}. Note that we anchor the past part of reparameterized $\mathbfz_s$ as the fixed condition in every step of the inference process following Gong \etal \cite{gong2022diffuseq}. Finally, the generated $\hat{F}^\smallR_\text{seq}$ and $\hat{F}^\smallL_\text{seq}$ are used to predict future hand waypoints and contact points by $f_{\scriptscriptstyle\text{HTH}}(\cdot)$ and $f_{\scriptscriptstyle\text{OAH}}(\cdot)$ as mentioned before. 
It can be seen from the inference stage depicted in Fig.~\ref{fig:compare_ar} that Diff-IP2D is iter-NAR with bidirectional constraints in the latent feature space (also refer to Sec.~2 of the supplementary material).

\begin{figure}[t]
  \centering
  \includegraphics[width=1\linewidth]{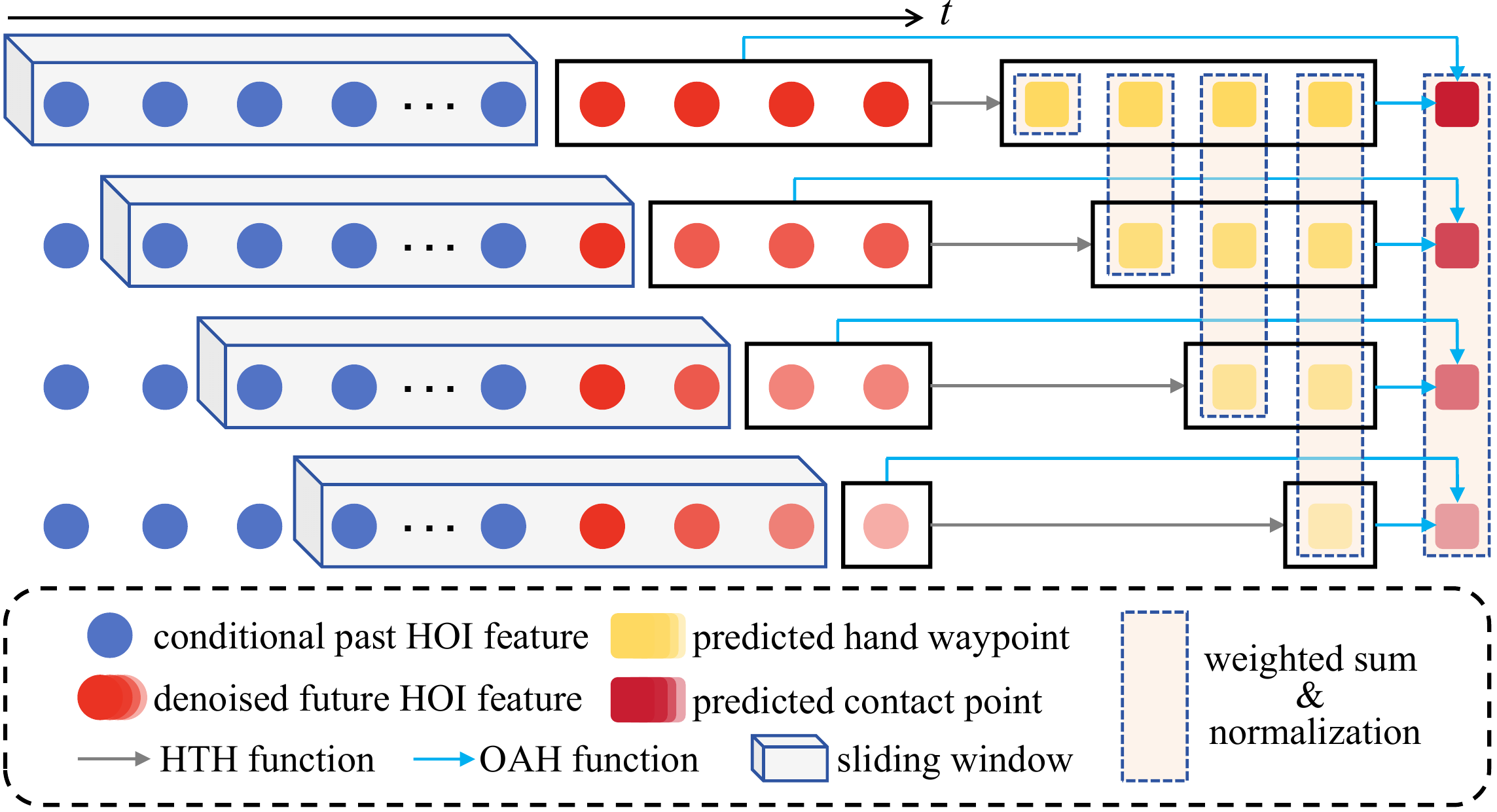}
  \caption{Temporal enhancement aggregates multiple prediction results to enhance forward constraints while still keeping backward constraints within each denoising process.}
  \label{fig:temp_enhance}
  \vspace{-0.7cm}
\end{figure}

\textbf{Temporal enhancement.} We propose an optional temporal enhancement strategy to further improve the prediction performance of Diff-IP2D inspired by \cite{zhao2023learning}. As Fig.~\ref{fig:temp_enhance} illustrates, we can incorporate an additional autoregressive setup at the end of our proposed iter-NAR paradigm. This helps to improve the smoothness of predicted hand trajectories and simultaneously refine affordance prediction. Specifically, we exploit a sliding window to predict hand waypoints and contact points at each future timestamp. Then an exponential weighting scheme $w_t=\text{exp}(-u*t)$ is adopted to aggregate these results, where $u$ is constant. Our proposed temporal enhancement does not affect the concrete pattern of each prediction (e.g. the fixed number of input frames), and thus there is no need to integrate this step-wise operation in the training stage. We denote Diff-IP2D with temporal enhancement as Diff-IP2D$^{\dag}$, which is suited to applications with low real-time requirements since multiple denoising operations should be performed to obtain ultimate prediction results. Recall that Diff-IP2D follows the iter-NAR paradigm considering bidirectional constraints in each HOI process, and Diff-IP2D$^{\dag}$ still keeps bidirectional constraints during denoising operation at each timestamp. The difference is that Diff-IP2D$^{\dag}$ enhances forward constraints to improve trajectory smoothness and reduce affordance uncertainty by weighting step-wise denoised HOI.

\begin{table*}[t]
\scriptsize
\setlength{\tabcolsep}{11pt}
\center
\renewcommand\arraystretch{0.7}
\caption{Comparison of performance on hand trajectory and object affordance prediction}
\vspace{-0.2cm}
\begin{tabular}{l|ccc|ccc|ccc}
\toprule
\multicolumn{1}{l|}{\multirow{2}{*}{approach}}   & \multicolumn{3}{c|}{EK55} & \multicolumn{3}{c|}{EK100}  & \multicolumn{3}{c}{EG} \\ \cmidrule{2-10} 
\multicolumn{1}{c|}{}                                                                               & WDE\,$\downarrow$  &  & FDE\,$\downarrow$ & WDE\,$\downarrow$  &  & FDE\,$\downarrow$ & WDE\,$\downarrow$  &  & FDE\,$\downarrow$   \\ \cmidrule{1-10}                                                                   CVH & 0.636   &   & 0.315     &  0.658  &   & 0.329    & 0.689  &   & 0.343    \\
Seq2Seq \cite{sutskever2014sequence}  & 0.505  &    & 0.212      &  0.556   &  & 0.219    & 0.649 &    & 0.263   \\ 
FHOI \cite{liu2020forecasting}   & 0.589    &     & 0.307     & 0.550   &     & 0.274    & 0.557   &  & 0.268   \\ 
OCT \cite{liu2022joint} & 0.446   &    & 0.208     & 0.467   &    & 0.206    & 0.514    &   & 0.249   \\ 
USST \cite{bao2023uncertainty}  & 0.458   &   & 0.210           & 0.475   &    & 0.206  & 0.552 &   & 0.256   \\ 
Diff-IP2D (ours)  & \textbf{0.411}   &    & \textbf{0.181}     &  \textbf{0.407}  &  & \textbf{0.187}    &  \textbf{0.478}  &  & \textbf{0.211}          \\ \cmidrule{1-10}
\multicolumn{1}{c|}{}                                                                               & SIM\,$\uparrow$    & AUC-J\,$\uparrow$ & NSS\,$\uparrow$  & SIM\,$\uparrow$    & AUC-J\,$\uparrow$ & NSS\,$\uparrow$  & SIM\,$\uparrow$    & AUC-J\,$\uparrow$ & NSS\,$\uparrow$   \\ \cmidrule{1-10} 
                             Center Object \cite{liu2020forecasting} & 0.083      & 0.553      & 0.448     &  0.081     & 0.558    & 0.401    & 0.094      & 0.562     & 0.518    \\
Hotspots \cite{nagarajan2019grounded}  & 0.156      &  0.670     & 0.606     &  0.147     & 0.635     & 0.533    & 0.150      & 0.662    & 0.574   \\ 
FHOI \cite{liu2020forecasting}   & 0.159      & 0.655      & 0.517     & 0.120      & 0.548    & 0.418    & 0.122      & 0.506    & 0.401  \\ 
OCT \cite{liu2022joint} & 0.213      & 0.710      & 0.791     & 0.187      & 0.677    & 0.695    & 0.227      & 0.704     & 0.912    \\ 
USST-FH \cite{bao2023uncertainty}  & 0.208      & 0.682     & 0.757     & 0.179        & 0.658    & 0.754    & 0.190      & 0.675    & 0.729   \\ 
Diff-IP2D (ours)  & \textbf{0.226}      & \textbf{0.725}      & \textbf{0.980}     &  \textbf{0.211}     & \textbf{0.736}    & \textbf{0.917}    & \textbf{0.242}      & \textbf{0.722}    & \textbf{0.956}        \\ \cmidrule{1-10}
& SIM$^*\!\uparrow$    & AUC-J$^{*\!}\uparrow$ & NSS$^{*\!}\uparrow$  & SIM$^{*\!}\uparrow$    & AUC-J$^{*\!}\uparrow$ & NSS$^{*\!}\uparrow$  & SIM$^{*\!}\uparrow$    & AUC-J$^{*\!}\uparrow$ & NSS$^{*\!}\uparrow$   \\ \cmidrule{1-10}                                
FHOI \cite{liu2020forecasting}   & 0.130      & 0.602      & 0.487     & 0.113      & 0.545    & 0.409    & 0.118      & 0.501    & 0.379  \\ 
OCT \cite{liu2022joint} & 0.219      & 0.720      & 0.848     & 0.182      & 0.684    & 0.662    & 0.194      & 0.672     & 0.752    \\ 
Diff-IP2D (ours)  & \textbf{0.222}      & \textbf{0.730}      & \textbf{0.888}     &  \textbf{0.204}     & \textbf{0.727}    & \textbf{0.844}    & \textbf{0.226}      & \textbf{0.701}    & \textbf{0.825}        \\ \bottomrule
\end{tabular}
\label{tab:compare_hand}
\vspace{-0.2cm}
\end{table*}

\begin{table*}[t] \scriptsize
\setlength{\tabcolsep}{8.5pt}
\center
\renewcommand\arraystretch{0.7}
\caption{Ablation study on egomotion guidance}
\vspace{-0.2cm}
\begin{tabular}{l|ccccc|ccccc}
\toprule
\multirow{2}{*}{approach}  & \multicolumn{5}{c|} {EK55} & \multicolumn{5}{c} {EK100}     \\ \cmidrule{2-11} 
      & WDE\,$\downarrow$ & FDE\,$\downarrow$ & SIM\,$\uparrow$ & AUC-J\,$\uparrow$      &NSS\,$\uparrow$  & WDE\,$\downarrow$ & FDE\,$\downarrow$ & SIM\,$\uparrow$ & AUC-J\,$\uparrow$      &NSS\,$\uparrow$  \\ \cmidrule{1-11} 
Diff-IP2D w/o egomotion guidance  & 0.427 & 0.186 & 0.218 & 0.717 & 0.929 & 0.439        & 0.198       & 0.201        & 0.710       & 0.846   \\
Diff-IP2D  & \textbf{0.411}       & \textbf{0.181}  & \textbf{0.226}      & \textbf{0.725}      & \textbf{0.980}  &\textbf{0.407}       & \textbf{0.187}       & \textbf{0.211}       & \textbf{0.736}       & \textbf{0.917} \\ \cmidrule{1-11}
improvement           & 3.7\% & 2.7\% & 3.7\% & 1.1\% & 5.5\%         &  7.3\%       & 5.6\%      &5.0\%       & 3.7\%       & 8.4\%          \\ \bottomrule  
\addlinespace[1pt]
\end{tabular}
\label{tab:ab_motion}
\vspace{-0.4cm}
\end{table*}

\vspace{-0.2cm}
\section{Experiments}
\label{sec:exp}

\subsection{Experimental Setups}
\label{sec:setups}

\textbf{Datasets.} Following the previous work \cite{liu2022joint,bao2023uncertainty}, we utilize four public datasets, Epic-Kitchens-55 (EK55) \cite{damen2018scaling}, Epic-Kitchens-100 (EK100) \cite{damen2022rescaling}, EGTEA Gaze+ (EG) \cite{li2018eye}, and EgoPAT3D-DT~\cite{{li2022egocentric,bao2023uncertainty}}. For the EK55 and EK100 datasets, we sample past $N_\text{p}=10$ frames (2.5\,s) to forecast HOI states in future $N_\text{f}=4$ frames (1.0\,s), both at 4 FPS. 
As to the EG dataset, $N_\text{p}=9$ frames (1.5\,s) are used for $N_\text{f}=3$ HOI predictions (0.5\,s) at 6 FPS. All the training and test splits of EK55, EK100, and EG are obtained following~\cite{liu2022joint}. More details can be found in Sec.~4 of the supplementary material. The experiments conducted on EgoPAT3D-DT are also provided in Sec.~5-D of the supplementary material.

\textbf{Diff-IP2D configuration.} MFE extracts the hand, object, and global feature vectors all with the size of $512$ for each input image. For the EK55 and EK100 datasets, the outputs of SOFM $F^\smallR_{\text{seq}}$, $F^\smallL_{\text{seq}}$ have the size of ${14\times 512}$ for training and ${10\times 512}$ for inference. For the EG dataset, $F^\smallR_{\text{seq}}$, $F^\smallL_{\text{seq}}$ are ${9\times 512}$ for training and ${12\times 512}$ for inference. As to the diffusion process, the total number of steps $S$ is set to $1000$. 
The square-root noise schedule in Diffusion-LM \cite{li2022diffusion} is adopted here for the forward diffusion process.
MADT has $6$ Transformer layers (Fig.~\ref{fig:transformer}) for denoising, where the embedding dimension is $512$, the number of heads is set to $4$, and the intermediate dimension of the feed-forward layer is set to $2048$.
Motion Encoder linearly projects each homography matrix to an egomotion feature vector of $512$. We use an MLP with hidden dimensions $256$ and $64$ to predict the hand waypoints as HTH, and a C-VAE containing an MLP with a hidden dimension $512$ to predict contact points as OAH. 
For training Diff-IP2D, we use AdamW optimizer~\cite{kingma2014adam} with a learning rate 2e-4. All the modules are trained for 30 epochs with a batch size of 8 on 2 A100 GPUs.
In the reference stage, we generate $10$ candidate samples for each prediction.

\textbf{Baseline configuration.} We choose Constant Velocity Hand (CVH), Seq2Seq \cite{sutskever2014sequence}, FHOI \cite{liu2020forecasting}, OCT \cite{liu2022joint}, and USST \cite{bao2023uncertainty} as the baselines for hand trajectory prediction. CVH is the most straightforward one, which assumes two hands remain in uniform motion over the future time horizon with the average velocity during past observations. 
We choose Center Object \cite{liu2020forecasting}, Hotspots \cite{nagarajan2019grounded}, FHOI \cite{liu2020forecasting}, OCT \cite{liu2022joint}, and Final Hand of USST \cite{bao2023uncertainty} (USST-FH) as the baselines for object affordance prediction. USST-FH puts a mixture of Gaussians at the last hand waypoint predicted by USST since its vanilla version can only predict waypoints.

\textbf{Evaluation metrics.} Following previous works \cite{liu2020forecasting,liu2022joint,bao2023uncertainty}, we use Final Displacement Error (FDE) to evaluate hand trajectory prediction performance. Considering the general knowledge of ``post-contact trajectories'' extracted from human videos is potentially beneficial to robot manipulation \cite{bahl2023affordances,mendonca2023structured}, we additionally extend the metric Average Displacement Error to Weighted Displacement Error (WDE):
\begin{align}
    \textstyle
    \text{WDE}=\frac{1}{2N_\text{f}}\sum_{\smallR,\smallL}\sum_{t=1}^{N_\text{f}}\frac{t}{N_\text{f}}D(H_t,H_t^\text{gt}),
\label{eq:wde}
\end{align}
where $D(\cdot)$ denotes the L2 distance function and later waypoints contribute to larger errors. We select the mean error among the 10 candidate samples for each trajectory prediction. As to object affordance prediction, we use Similarity Metric (SIM) \cite{swain1991color}, AUC-Judd (AUC-J) \cite{judd2009learning}, and Normalized Scanpath Saliency (NSS) \cite{peters2005components} as evaluation metrics. We use all 10 contact point candidates to compute them.

Moreover, we exploit an object-centric protocol to jointly evaluate the two prediction tasks. We first calculate the averaged hand waypoints $\bar{H}^\smallR_t$ and $\bar{H}^\smallL_t$ for each future timestamp from multiple samples. Then we select the waypoint closest to each predicted contact prediction $O_n$ as an additional possible contact point.
The joint hotspot is predicted with the additional contact points and $O_n$. This comprehensively considers object-centric attention since HOI changes object states and hand waypoints must have a strong correlation with object positions. Here we use the quantitative metrics same as the ones for affordance prediction, denoted as SIM$^{*}$, AUC-J$^{*}$, and NSS$^{*}$.

\begin{figure}
  \centering
  \captionsetup{aboveskip=2pt, belowskip=0pt}
  \includegraphics[width=1\linewidth]{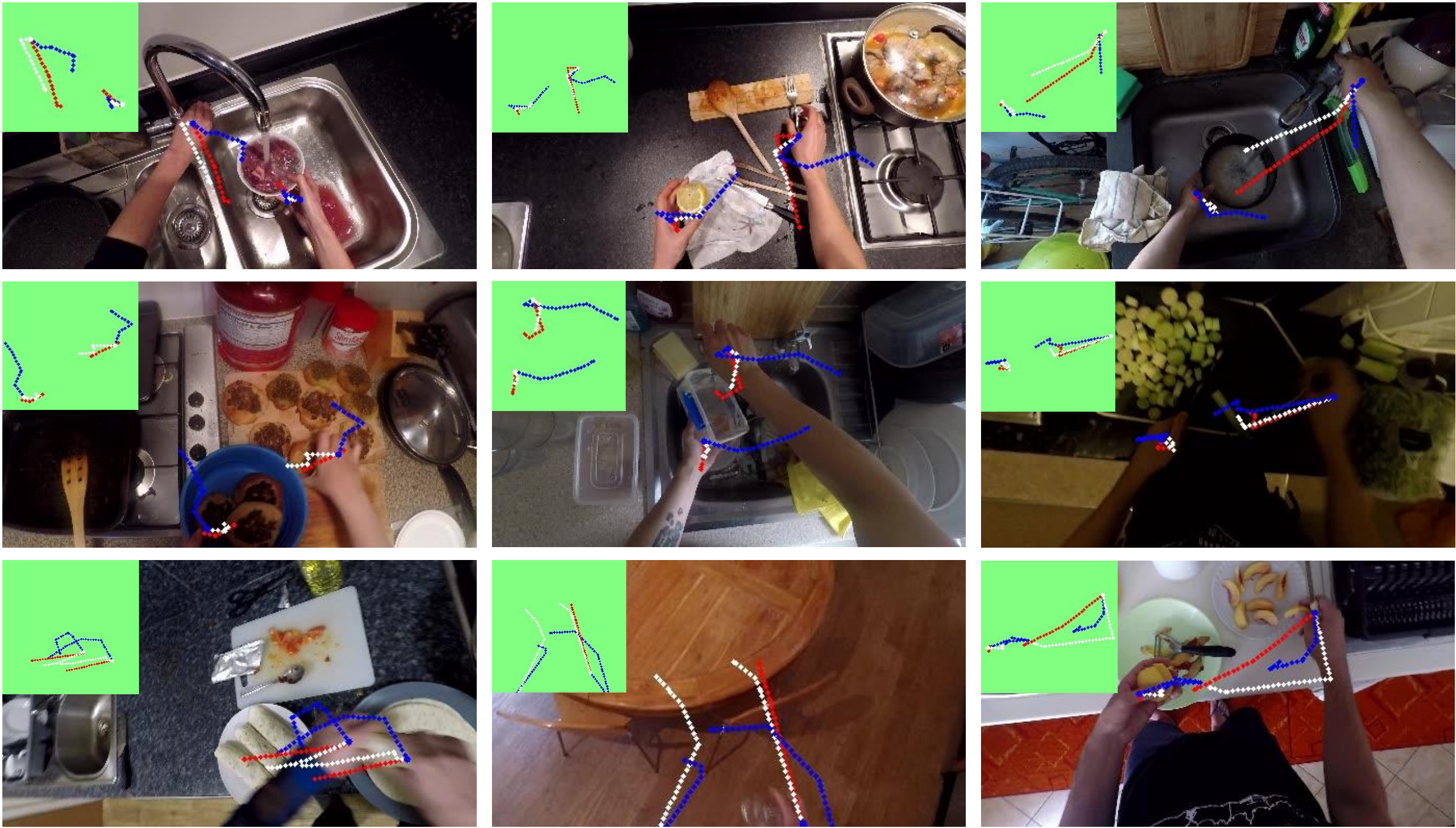}
  \caption{Visualization of hand trajectory prediction on Epic-Kitchens. Waypoints from GT, Diff-IP2D, and the second-best baseline \cite{liu2022joint} are connected by red, white, and blue dashed lines.}
  \label{fig:hand_pred}
  \vspace{-0.7cm}
\end{figure}

\begin{figure}
  \centering
  \captionsetup{aboveskip=2pt, belowskip=0pt}
  \includegraphics[width=1\linewidth]{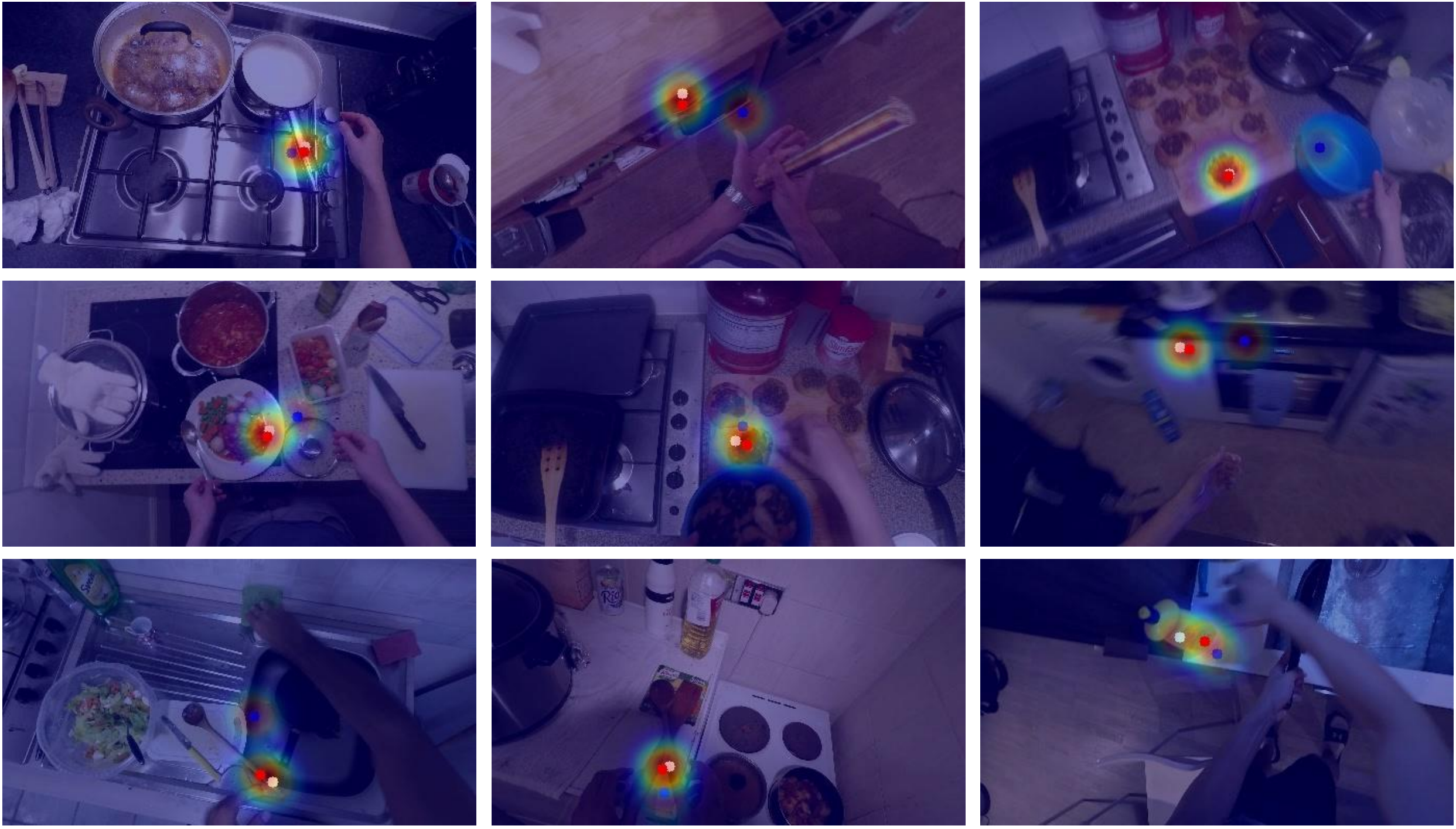}
  \caption{Visualization of object affordance prediction on Epic-Kitchens. Contact points from GT, Diff-IP2D, and the SOTA baseline OCT \cite{liu2022joint} are represented by red, white, and blue dots respectively. For clearer illustration, we additionally put a fixed Gaussian with each contact point as the center.}
  \label{fig:obj_pred}
  \vspace{-0.7cm}
\end{figure}

\vspace{-0.1cm}
\subsection{Separate Evaluation on Hand Trajectory and Object Affordance Prediction}
\label{sec:separate_eval}
We first present the evaluation results on hand trajectory prediction. 
As Tab.~\ref{tab:compare_hand} depicts, our proposed Diff-IP2D outperforms all the baselines on both EK55 and EK100 on WDE and FDE. This is mainly achieved by the devised iter-NAR paradigm of Diff-IP2D alleviating degeneration in AR baselines, as well as the egomotion guidance. The visualization of hand prediction results in Fig.~\ref{fig:hand_pred} shows that Diff-IP2D can better capture the camera wearer's intention (such as putting the food in the bowl) and generate more reasonable future trajectories even if lacking past observations for hands (such as reaching out towards the table). Besides, Diff-IP2D can predict a good final hand position despite a large shift in the early stage (Fig.~\ref{fig:hand_pred}, bottom right), owing to our parallel generation with bidirectional constraints. When directly transferring the models trained on Epic-Kitchens to the unseen EG dataset, Diff-IP2D still improves the second-best baselines by 7.0\% and 15.3\% on WDE and FDE respectively. This reveals the solid generalization capability of our Diff-IP2D across different environments.

The comparison results of object affordance prediction are also shown in Tab.~\ref{tab:compare_hand}. Our proposed Diff-IP2D predicts the hotspots with larger SIM, AUC-J, and NSS compared to all the baselines on both Epic-Kitchens data and unseen EG data.
Fig.~\ref{fig:obj_pred} illustrates the predicted contact points with minimum distances to the ground-truth ones. Our proposed method focuses more on objects of interest considering the features of the holistic interaction and potential hand trajectories, and therefore grounds the contact points closer to the ground-truth labels than the counterparts of the baseline.

\begin{table*}[t] \scriptsize
\setlength{\tabcolsep}{14pt}
\center
\renewcommand\arraystretch{0.7}
\caption{Ablation study on supervision signals}
\vspace{-0.2cm}
\begin{tabular}{l|cccccccc}
\toprule
\multirow{2}{*}{approach}  & \multicolumn{2}{c} {hand trajectory}  &\multicolumn{3}{c} {object affordance}   &\multicolumn{3}{c} {joint evaluation}     \\ \cmidrule{2-9} 
      & WDE\,$\downarrow$ & FDE\,$\downarrow$ & SIM\,$\uparrow$ & AUC-J\,$\uparrow$      &NSS\,$\uparrow$  & SIM$^{*\!}\uparrow$ & AUC-J$^{*\!}\uparrow$      &NSS$^{*\!}\uparrow$  \\ \cmidrule{1-9} 
Diff-IP2D w/o diff.   & 0.480        & 0.201       & 0.142        & 0.624       & 0.406  & 0.189  & 0.634  & 0.764  \\
Diff-IP2D w/o reg.   & 0.430        & 0.195       & 0.205        & 0.718       & 0.821 & 0.180  & 0.692  & 0.722  \\
Diff-IP2D    &\textbf{0.407}       & \textbf{0.187}       & \textbf{0.211}       & \textbf{0.736}       & \textbf{0.917} 
 & \textbf{0.204}    & \textbf{0.727}   & \textbf{0.844}   \\ \bottomrule
\end{tabular}
\label{tab:ab_reg}
\vspace{-0.2cm}
\end{table*}

\begin{table*}[t] \scriptsize
\setlength{\tabcolsep}{11pt}
\center
\renewcommand\arraystretch{0.7}
\caption{Diff-IP2D vs. Diff-IP2D$^{\dag}$ with temporal enhancement}
\vspace{-0.2cm}
\begin{tabular}{l|ccccc|ccccc}
\toprule
\multirow{2}{*}{approach}  & \multicolumn{5}{c|} {EK55} & \multicolumn{5}{c} {EK100}     \\ \cmidrule{2-11} 
      & WDE\,$\downarrow$ & FDE\,$\downarrow$ & SIM\,$\uparrow$ & AUC-J\,$\uparrow$      &NSS\,$\uparrow$  & WDE\,$\downarrow$ & FDE\,$\downarrow$ & SIM\,$\uparrow$ & AUC-J\,$\uparrow$      &NSS\,$\uparrow$  \\ \cmidrule{1-11} 
Diff-IP2D  & {0.411}       & {0.181}  & {0.226}      & {0.725}      & {0.980}  &{0.407}       & {0.187}       & {0.211}       & {0.736}       & {0.917} \\
Diff-IP2D$^{\dag}$  & \textbf{0.388} & \textbf{0.172} & \textbf{0.230} & \textbf{0.735} & \textbf{0.992} & \textbf{0.395}        & \textbf{0.179}       & \textbf{0.217}        & \textbf{0.744}       & \textbf{0.930}   \\ \cmidrule{1-11}
improvement           & 5.6\% & 5.0\% & 1.8\% & 1.4\% & 1.2\%         &  2.9\%       & 4.3\%      & 2.8\%       & 1.1\%       & 1.4\%          \\ \bottomrule  
\addlinespace[1pt]
\end{tabular}
\label{tab:ab_temporal_enhance}
\vspace{-0.3cm}
\end{table*}

\subsection{Joint Evaluation on HOI Prediction}
\label{sec:joint_eval}
We further compare Diff-IP2D with the other two joint prediction baselines, FHOI \cite{liu2020forecasting} and OCT \cite{liu2022joint}, using the object-centric protocol. The video clips containing both ground-truth hand waypoints and contact points are used for evaluation in this experiment. The results are also shown in Tab.~\ref{tab:compare_hand}, which indicates that our proposed Diff-IP2D can generate the best object-centric HOI predictions considering the two tasks concurrently on both Epic-Kitchens and unseen EG data. The results also suggest that Diff-IP2D outperforms the baselines on object-centric HOI prediction by focusing more attention on the target objects and predicting reasonable hand trajectories around them.

\begin{table}[t] \scriptsize
\setlength{\tabcolsep}{10pt}
\center
\renewcommand\arraystretch{1.1}
\caption{Ablation study on the denoising model}
\vspace{-0.2cm}
\begin{tabular}{l|cc|cc}
\toprule
\multirow{2}{*}{denoising}   & \multicolumn{2}{c|} {EK55} & \multicolumn{2}{c} {EK100} \\ \cmidrule{2-5} 

&WDE\,$\downarrow$ & FDE\,$\downarrow$  &WDE\,$\downarrow$ & FDE\,$\downarrow$     \\ \cmidrule{1-5} 
separate models   & 0.425    & 0.189   & 0.412  & 0.189             \\
unified model  & \textbf{0.411}    & \textbf{0.181}   & \textbf{0.407}   & \textbf{0.187}       \\ \bottomrule
\end{tabular}\\
\label{tab:abla_denoising_model}
\vspace{-0.5cm}
\end{table}

\subsection{Ablation Studies}
\textbf{Egomotion guidance.} We first ablate the egomotion features used to guide MADT denoising on the EK55 and EK100 datasets. Here we replace the MHCA in MADT with a multi-head self-attention module (MHSA) to remove the egomotion guidance while keeping the same parameter number. The experimental results in Tab.~\ref{tab:ab_motion} show that the guidance of motion features improves our proposed diffusion-based paradigm noticeably on both hand trajectory prediction and object affordance prediction. This is achieved by narrowing the two gaps caused by 2D-3D ill-posed problem and view difference mentioned in Sec.~\ref{sec:intro}. Note that the egomotion guidance is more significant on the EK100 dataset than on the EK55 dataset. The reason could be that EK100 has a larger volume of training data incorporating more diverse egomotion patterns than EK55, leading to a model that can capture human dynamics better.

\textbf{Supervision signals.} We provide an additional ablation study on diffusion-related losses, and the regularization term which links $\{F^\smallR_{\text{seq}},F^\smallL_{\text{seq}}\}$ and $\{\hat{F}^\smallR_{\text{seq}},\hat{F}^\smallL_{\text{seq}}\}$. The experimental results on the EK100 dataset are shown in Tab.~\ref{tab:ab_reg}. Diff-IP2D without diffusion-related losses witness significant performance degradation due to a lack of dense supervision signals in the latent space beyond sparse constraints in the image plane. We thus argue that high-level dense supervision aligns the model with high-level human intentions for better prediction performance. Tab.~\ref{tab:ab_reg} also shows that our regularization strategy remarkably enhances prediction performance on both hand trajectories and object affordances even if it is only used to link the latent space with hand trajectory prediction. More analysis of the regularization can be found in Sec.~3 of the supplementary material.

\textbf{Denoising model.} We conduct a baseline by instantiating two separate MADT models for right and left sides, and denoise side-oriented HOI features of two sides respectively. Tab.~\ref{tab:abla_denoising_model} shows that denoising with a unified MADT for both sides outperforms separate denoising. This suggests that side-oriented denoising with a unified model rather than separate ones helps to capture potential correlations between right and left sides, leading to higher prediction accuracy.

\textbf{Temporal enhancement.} We ablate our proposed temporal enhancement by comparing our vanilla Diff-IP2D to Diff-IP2D$^{\dag}$ mentioned in Sec.~\ref{sec:inference} on EK55 and EK100. Tab.~\ref{tab:ab_temporal_enhance} shows that the proposed temporal enhancement improves Diff-IP2D prediction accuracy. Note that temporal enhancement is an optional strategy since Diff-IP2D has already outperformed baselines. We advocate using Diff-IP2D$^{\dag}$ when there is no requirement of high efficiency for better HOI prediction performance.

\subsection{Key Findings} 
\label{sec:takeaways}
\textbf{Iter-NAR paradigm.} Compared to the SOTA baselines in an autoregressive manner shown in Tab.~\ref{tab:compare_hand}, Diff-IP2D shifts limited iterations along the time axis to sufficient iterations in the diffusion denoising direction, as Fig.~\ref{fig:compare_ar} shows. This alleviates accumulated artifacts caused by the limited iterations with forward-only constraints in the time dimension, and maintains bidirectional constraints (inherent in MADT) among sequential features to generate future HOI states in parallel. Bidirectional constraints respect \textit{spatial causality} where possible final HOI states also affect prior hand trajectories, without losing \textit{temporal causality}. We argue that this new paradigm provides a deeper understanding of high-level human intention for more accurate HOI prediction. 

\textbf{Concurrent motion capture.} Diff-IP2D with egomotion guidance is inherently suited to egocentric views because it concurrently captures hand/object movements and the camera wearer's egomotion patterns (homography) by the proposed MADT. It respects the fact that the changes of hand/object locations within the field of vision are entangled with human head motion following specific intentions in different activities. Tab.~\ref{tab:ab_motion} shows how egomotion affects HOI prediction positively. Additionally, our paradigm may also be beneficial to HOI prediction in non-egocentric views. For instance, if the human body egomotion can be captured in a third-person perspective, our model can also associate it with hand movements, thereby achieving better HOI prediction. We will explore this topic in future work.

\textbf{Side-oriented denoising with a unified model.} Diff-IP2D uses SOFM to separately fuse and denoise features for the left and right sides. We argue that the optimization directions for left-hand and right-hand HOI are different considering their different motion patterns and approaches to active objects. Therefore, separate noise sampling should be applied for the respective side in the training process. Notably, as Tab.~\ref{tab:abla_denoising_model} presents, we advocate using a unified model MADT to denoise for both sides with different sampled noises since we encourage it to capture the potential correlation between sides during one interaction process. 

\textbf{Dense supervision signals.} We extend sparse supervision signals from explicit ground-truth hand trajectories and contact points to the latent space. Specifically, we incorporate reconstructed implicit HOI features into training losses in our diffusion-based scheme, alongside explicit ground-truth supervision. We argue that this high-level dense supervision aligns the model with high-level human intentions, leading to better HOI prediction as Tab.~\ref{tab:ab_reg} shows.

\vspace{-0.1cm}
\section{Conclusion}
\vspace{-0.1cm}
\label{sec:discuss}
In this paper, we propose a novel hand-object interaction prediction method Diff-IP2D. It implements the devised denoising diffusion in the latent space under our proposed egomotion guidance, and jointly predicts future hand trajectories and object affordances with recovered latents on 2D egocentric videos. Experimental results validate that Diff-IP2D dominates the existing baselines on extended metrics, suggesting promising applications in artificial intelligence systems. 
We hope the takeaways about the iter-NAR paradigm, concurrent motion capture, side-oriented denoising with a unified model, and dense supervision signals in Diff-IP2D could inspire future HOI prediction research.

\bibliographystyle{ieeetr}

\footnotesize{
\bibliography{root}}

\newpage
\setcounter{section}{0}
\setcounter{figure}{0}
\setcounter{table}{0}
\renewcommand{\thesection}{\arabic{section}}
\renewcommand{\thefigure}{\arabic{figure}}
\renewcommand{\thetable}{\arabic{table}}
\renewcommand{\thesection}{\arabic{section}}

\begin{flushleft}
    \huge Supplementary Material
\end{flushleft}

\section{Motion-Related Gaps and Egomotion Homography}
\label{sec:app_motion}
In this section, we provide a detailed analysis for filling the motion-related gaps in hand-object interaction (HOI) prediction mentioned in Sec.~I of the main text with the egomotion homography. To narrow the view gap between the last observation and the other observations, homography works as a bridge to connect the pixel positions $\mathbf{p}_0$, $\mathbf{p}_t \in \mathbb{R}^{2}$ of one 3D hand waypoint/contact point on $I_t$ ($t\leq0$) and $I_0$, which can be represented by $\mathbf{p}_0=M_t\mathbf{p}_t$. We let the denoising network be aware of the egomotion features $E_t$ encoded from $M_t$ and enable it to capture the above-mentioned transformation when predicting future hand trajectories and contact points on the last observed image as a canvas.

For the 2D-3D gaps, we first discover the relationship between 2D pixel movements and 3D hand movements. For a 3D point that moves from $\mathbf{P}_t \in \mathbb{R}^3$ in the camera coordinate system at timestamp $t$ ($t\leq0$) to $\mathbf{P}_0 \in \mathbb{R}^3$ in the camera coordinate system at timestamp $t=0$, we first project them to the image plane by $\mathbf{p}_t=K\mathbf{P}_t$ and $\mathbf{p}_0=K\mathbf{P}_0$, where $K$ is the intrinsic parameters. Then we transform  $\mathbf{p}_t$ to the last canvas image by $\mathbf{p}'_t=M_t\mathbf{p}_t$. The 2D pixel movement on the last image can be formulated as:
\begin{align}
\mathbf{p}_0 - \mathbf{p}'_t = K\mathbf{P}_0 - M_t\mathbf{p}_t = K\mathbf{P}_0 - M_tK\mathbf{P}_t. \nonumber
\end{align}
Therefore, the 3D action ($\mathbf{P}_t\rightarrow \mathbf{P}_0$) uniquely corresponds to the 2D pixel movement ($\mathbf{p}_t\rightarrow \mathbf{p}_0$) once $K$ and $M_t$ are both determined. Since $K$ is a constant for each video clip, only $M_t$ changing along the time axis determines the spatial relationship between observations. Therefore, we enable our proposed model aware of egomotion by encoding $M_t$ to a feature vector absorbed by multi-head cross attention of Motion-Aware Denoising Transformer as mentioned in Sec.~III-B of the main text, narrowing the gap between 2D pixel movement and 3D actions. Note that we do not utilize SE(3) here due to scale-agnostic estimation with only 2D images as input.

\section{Iterative Non-Autoregressive Paradigm vs. Autoregressive Paradigm}

\begin{figure*}[t]
  \centering
  \captionsetup{aboveskip=2pt, belowskip=0pt}
  \includegraphics[width=1\linewidth]{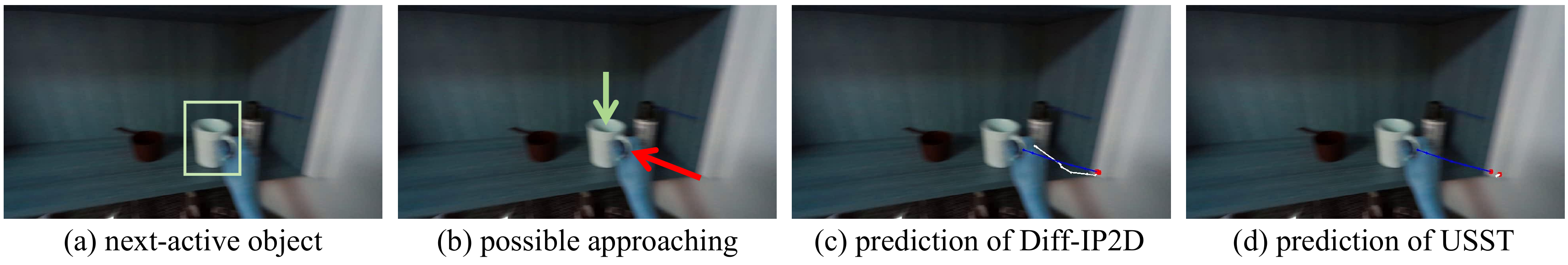}
  \caption{An example to clarify our motivation to propose an iter-NAR paradigm considering bidirectional constraints. The hand waypoints from ground-truth labels and HOI prediction approaches are connected by blue and white dashed lines respectively. Note that we reverse the RGB values of each image to display the arm’s positions more clearly. There is a lack of backward constraints in AR-based USST \cite{bao2023uncertainty}, leading to a shorter predicted trajectory (almost curled up into a point) and larger accumulated displacement errors. In contrast, our Diff-IP2D with iter-NAR paradigm is potentially guided by final HOI states, and thus predicts more accurate hand trajectories following both spatial causality and temporal causality.}
  \label{fig:motivation}
  \vspace{-0.3cm}
\end{figure*}

Our proposed Diff-IP2D is an iterative non-autoregressive (iter-NAR) model, showing better HOI prediction performance compared to the state-of-the-art methods~\cite{liu2022joint,bao2023uncertainty} with the autoregressive (AR) paradigm.
AR models reason about the next HOI state only according to the previous steps (Fig. I(a) in the main text), leading to the forward-only constraint. They overlook the backward constraint which we think is also important for HOI prediction. 
We provide an example in Fig.~\ref{fig:motivation} to further explain the significance of the backward constraint. The human hand generally picks up a cup (Fig.~\ref{fig:motivation}(a)) by its handle because side-gripping by the handle is more stable and allows for a faster target approach than other ways. It is more likely for a hand to approach the cup from the side (red arrow in Fig.~\ref{fig:motivation}(b)) than from the top (green arrow in Fig.~\ref{fig:motivation}(b)) in the near future. Consequently, the final state of the future HOI can be approximately determined, which dictates the hand movement toward the cup, thereby establishing potential backward constraints on \textit{spatial causality}. Therefore, we argue that HOI prediction should be modeled as the non-autoregressive process considering the bidirectional constraints within the holistic sequence, rather than the autoregressive process with only forward constraints on \textit{temporal causality}. 

\begin{figure}[t]
  \centering
  \captionsetup{aboveskip=2pt, belowskip=0pt}
  \includegraphics[width=0.6\linewidth]{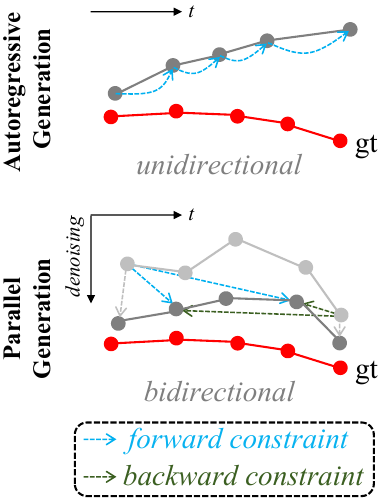}
  \caption{Autoregressive generation vs. parallel generation.}
  \label{fig:nar_vs_ar}
  \vspace{-0.3cm}
\end{figure}

\begin{table*}[t] \small
\setlength{\tabcolsep}{19pt}
\center
\renewcommand\arraystretch{1.1}
\caption{Joint evaluation results in the ablation study on egomotion guidance}
\begin{tabular}{l|ccc|ccc}
\toprule
\multirow{2}{*}{approach}  & \multicolumn{3}{c|} {EK55} & \multicolumn{3}{c} {EK100}     \\ \cmidrule{2-7} 
      & SIM$^{*\!}\uparrow$ & AUC-J$^{*\!}\uparrow$      &NSS$^{*\!}\uparrow$  & SIM$^{*\!}\uparrow$ & AUC-J$^{*\!}\uparrow$      &NSS$^{*\!}\uparrow$  \\ \cmidrule{1-7} 
Diff-IP2D*   & 0.216   & 0.718   & 0.842        & 0.198        & 0.712       & 0.778   \\
 Diff-IP2D        & \textbf{0.222}      & \textbf{0.730}      & \textbf{0.888}     &  \textbf{0.204}     & \textbf{0.727}    & \textbf{0.844} \\ \cmidrule{1-7}
improvement          & 2.8\% & 1.7\% & 5.5\%  & 3.0\%       & 2.1\%    & 8.5\%      \\ \bottomrule  
\multicolumn{7}{l}{\scriptsize Diff-IP2D*: Diff-IP2D w/o egomotion guidance}
\end{tabular}
\label{tab:ab_motion_supp}
\vspace{-0.4cm}
\end{table*}

We also provide an illustration comparison between iter-NAR parallel generation and AR generation in Fig.~\ref{fig:nar_vs_ar}. Our proposed iter-NAR paradigm predicts future HOI states in parallel considering bidrectional constraints encompassing both forward and backward constraints within the holistic interaction sequence. It also shifts the limited iterations along the time axis to the sufficient iterations in the diffusion denoising direction (also shown in Fig.~IV of the main text). 
Following the derivation of the previous work DiffuSeq \cite{gong2022diffuseq} which is used for text generation, here we further mathematically prove that our proposed Diff-IP2D prediction process can be regarded as an iter-NAR process. We first introduce a series of intermediate HOI states $\{\mathbf{F}_s^y\}_{s=0}^{S}$ decoded from $\{\mathbf{y}_s\}_{s=0}^{S}$, where $\mathbf{y}_s$ denotes the future part of $\mathbfz_s$ and $\mathbf{y}_S \sim \mathcal{N}(0,\textbf{\text{I}})$. 
$\mathbf{F}^x$ represents the past latent HOI features $F_{\text{seq}}^\smallR$ or $F_{\text{seq}}^\smallL$ from Side-Oriented Fusion Module. $\mathbf{M}$ denotes the egomotion guidance $M_\text{seq}$ here and will be extended by other perception information in our future work. Therefore, the inference process of our proposed diffusion-based approach can be formulated as follows:
\tinytext{
\begin{minipage}{\linewidth}
\begin{align}
&~p_{\scriptscriptstyle\text{Diff-IP2D}}(\mathbf{F}^y|\mathbf{F}^x)  \nonumber \\
=& {\sum_{\mathbf{F}^y_S,\ldots,\mathbf{F}^y_1}\int_{\mathbf{y}_{S},\ldots,\mathbf{y}_0}}{p(\mathbf{F}^y|\mathbf{y}_{0},\mathbf{F}^x)}\prod_{s=S,\ldots,1}{p(\mathbf{y}_{s-1}|\mathbf{F}^y_s)p(\mathbf{F}^y_{s}|\mathbf{y}_s,\mathbf{F}^x,\mathbf{M})} \nonumber \\ 
=& {\sum_{\mathbf{F}^y_S,\ldots,\mathbf{F}^y_1}\int_{\mathbf{y}_{S},\ldots,\mathbf{y}_0}}p(\mathbf{F}^y_S|\mathbf{y}_S,\mathbf{F}^x)\prod_{s=S-1,\ldots,0}{p(\mathbf{F}^y_s|\mathbf{y}_s,\mathbf{F}^x,\mathbf{M})p(\mathbf{y}_s|\mathbf{F}^y_{s+1})} \nonumber \\ 
=&\sum_{\substack{\mathbf{F}^y_S,\ldots,\mathbf{F}^y_1}}p(\mathbf{F}^y_S|\mathbf{y}_S,\mathbf{F}^x)\prod_{s=S-1,\ldots,0}{ {\int}_{\mathbf{y}_s}p(\mathbf{F}^y_s|\mathbf{y}_s,\mathbf{F}^x,\mathbf{M})p(\mathbf{y}_s|\mathbf{F}^y_{s+1})}.  \nonumber
\end{align}
\end{minipage}}

\noindent Then we marginalize over $\mathbf{y}$ and obtain the initial iterative non-autoregressive form of our proposed approach:
\tinytext{
\begin{minipage}{\linewidth}
\begin{align}
&~p_{\scriptscriptstyle\text{Diff-IP2D}}(\mathbf{F}^y|\mathbf{F}^x) \nonumber \\ 
=&\sum_{\substack{\mathbf{F}^y_S,\ldots,\mathbf{F}^y_1}}p(\mathbf{F}^y_S|\mathbf{y}_S,\mathbf{F}^x)\prod_{t=S-1,\ldots,0}{p(\mathbf{F}^y_s|\mathbf{F}^y_{s+1},\mathbf{F}^x,\mathbf{M})} \nonumber \\ 
\equiv&\sum_{\mathbf{F}^y_1,\ldots,\mathbf{F}^y_{K-1}}{p(\mathbf{F}^y_1|\mathbf{F}^x)\vphantom{\prod_{k=1,\ldots,K-1}}\prod_{k=1,\ldots,K-1}{p(\mathbf{F}^y_{k+1}|\mathbf{F}^y_k,\mathbf{F}^x,\mathbf{M})}},  \nonumber 
\end{align}
\end{minipage}}

\noindent where we align the variable $s$, which denotes the diffusion steps, with the commonly used iteration variable $k$ in typical iterative formulas. Here what we pursue using the denoising diffusion model is to recover implicit features of future HOI states instead of directly decoding the final explicit hand waypoints or contact points. Therefore, we can regard the iterative process (latents$\rightarrow$ explicit HOI$\rightarrow$ latents) inherent in the above-mentioned equation as an equivariant mapping (latents$\rightarrow$ latents). The above equation can be further transformed to the ultimate iter-NAR form of our proposed Diff-IP2D:
\tinytinytext{
\begin{minipage}{\linewidth}
\begin{align}
&~p_{\scriptscriptstyle\text{Diff-IP2D}}(\mathbf{y}|\mathbf{F}^x) \nonumber \\ 
=&\sum_{\mathbf{y}_1,\ldots,\mathbf{y}_{K-1}}{p(\mathbf{y}_1|\mathbf{F}^x)\vphantom{\prod_{k=1,\ldots ,K-1}}\prod_{k=1,\ldots,K-1}{p(\mathbf{y}_{k+1}|\mathbf{y}_k,\mathbf{F}^x,\mathbf{M})}}\nonumber \\ 
=&\sum_{\mathbf{y}_1,\ldots,\mathbf{y}_{K-1}}{\prod_{i=1,\ldots,N_\text{f}}p(\mathbf{y}_{1,i}|\mathbf{F}^x)\vphantom{\prod_{k=1\ldots K-1}}\prod_{k=1,\ldots ,K-1}\prod_{i=1,\ldots,N_\text{f}}{p(\mathbf{y}_{k+1,i}|\mathbf{y}_{k,1:N_\text{f}},\mathbf{F}^x,\mathbf{M})}}.  \nonumber  \\   \nonumber  
\end{align}
\end{minipage}}

\section{Motivation of the Regularization Loss}
We propose a regularization term $\mathcal{L}_\text{reg}$ in Eq.~(4) of the main text for better model optimization. Here we provide more details about the motivation of the regularization loss. In the training process, the tokenizer embeds RGB information and hand-object locations to latent features for the following denoising diffusion process. Here we describe the detailed function of the tokenizer: For each input image, we first exploit a pretrained Temporal Segment Network~\cite{furnari2020rolling} and extract hand and object RoIAlign~\cite{he2017mask} features given the detected bounding boxes from \cite{shan2020understanding}. Specifically, the center coordinates of the detected bounding boxes are encoded into the hand and object intermediate features, meaning that the latent features transformed from them encompass the spatial information of hands and objects within each image. After being corrupted to noisy features in the forward process and being denoised in the reverse process, the reconstructed latents are further transformed into locations of future hands and contact points by the predictors, including Hand Trajectory Head and Object Affordance Head. As can be seen, the latents are generated from input HOI states by the tokenizer before the forward process, and are further converted to output HOI states by the predictor after the reverse process. This is why we regarded latents before and after the denoising diffusion process as representing the same ``profile'' of the input HOI sequence. They both inherently encompass HOI state information in the same interaction duration, and the training process can be further regarded as the predictor distilling HOI state knowledge from the tokenizer. Therefore, we build a closer gradient connection between the tokenizer and the predictor by introducing the regularization term into the training process to enhance the knowledge distillation. Tab.~III in the main text presents the improvement in HOI prediction from our proposed regularization strategy.

\section{More Details about Datasets and Diff-IP2D Training Configurations}
\label{sec:add_implement_detail}
The training sets of EK55~\cite{damen2018scaling} and EK100~\cite{damen2022rescaling} contain 8523 and 24148 video clips respectively. Their test sets consist of 1894 and 3513 samples for hand trajectory evaluation, and 241 and 401 samples for object affordance evaluation. In contrast to Epic-Kitchens, the EG dataset~\cite{li2018eye} offers a smaller data volume, including 1880 training samples, 442 evaluation hand trajectories, and 69 evaluation interaction hotspots. All the training sets are automatically generated following Liu \etal \cite{liu2022joint}. Note that we exclusively use the test part of the EG dataset to assess generalization ability in the experiments of Sec.~IV-B and Sec.~IV-C since it contains insufficient training samples for reasonable convergence. 

For training Diff-IP2D, we use AdamW optimizer \cite{kingma2014adam} with a learning rate $2$e-$4$. The total loss function is depicted below. The loss weights are initially set as $\lambda_{\text{VLB}}=1$, $\lambda_{\text{traj}}=1$, $\lambda_{\text{aff}}=0.1$, and $\lambda_{\text{reg}}=0.2$. All the networks in Diff-IP2D are trained for $30$ epochs with a batch size of $8$ on 2 A100 GPUs.
\begin{align}
    \mathcal{L}_\text{total} = 
    \lambda_{\text{VLB}}(\mathcal{L}_\text{VLB}^\smallR + \mathcal{L}_\text{VLB}^\smallL) + 
    \lambda_{\text{traj}}(\mathcal{L}_\text{traj}^\smallR + \mathcal{L}_\text{traj}^\smallL) \nonumber\\ + 
    \lambda_{\text{aff}}\mathcal{L}_\text{aff} + 
    \lambda_{\text{reg}}(\mathcal{L}_\text{reg}^\smallR + \mathcal{L}_\text{reg}^\smallL). \nonumber
\end{align}

\section{Additional Experimental Results}

\subsection{Joint Evaluation on the Effect of Egomotion Guidance}

We present the supplementary evaluation results in the ablation study on egomotion guidance. Our proposed joint evaluation protocol is applied here to show the positive effect of egomotion guidance for denoising diffusion. As can be seen in Tab.~\ref{tab:ab_motion_supp}, the use of the egomotion features enhances the joint prediction performance of Diff-IP2D on both EK55 and EK100. EK100 has a larger data volume which contains much more human motion patterns than EK55, leading to larger improvement on SIM$^*$, AUC-J$^*$, and NSS$^*$ by $3.0\%$, $2.1\%$, and $8.5\%$ respectively.

\subsection{Ablation Study on Observation Time}
\begin{figure}[t]
    \centering
    \begin{subfigure}[b]{1\linewidth}
        \includegraphics[width=\textwidth]{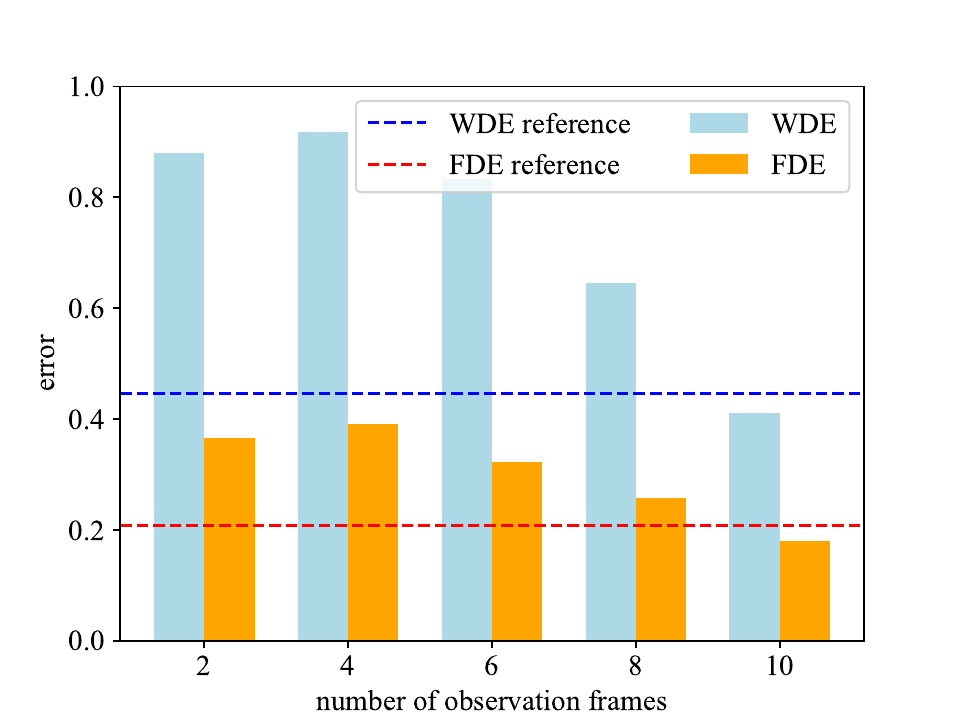}
        \caption{Zero-shot prediction}
        \label{fig:compare_obs_sub1}
    \end{subfigure}
    \hfill
    \begin{subfigure}[b]{1\linewidth}
        \includegraphics[width=\textwidth]{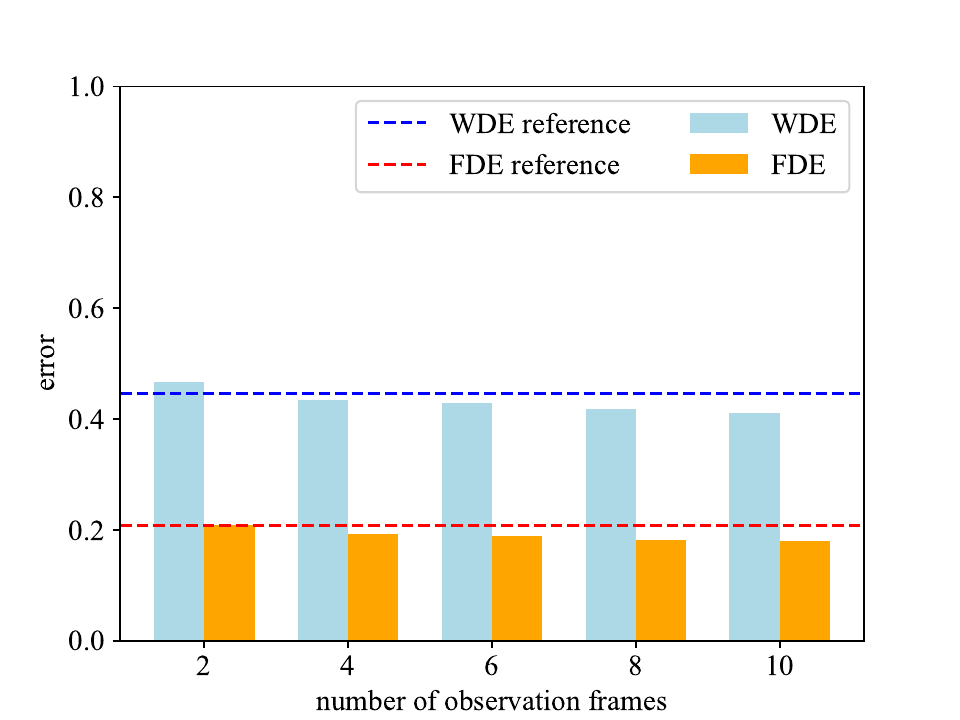}
        \caption{Prediction by models trained from scratch}
        \label{fig:compare_obs_sub2}
    \end{subfigure}
  \caption{Ablation study on observation time. The reference line represents the performance of the second-best baseline trained from scratch using $10$ observation frames.}
    \label{fig:compare_obs}
    \vspace{-0.5cm}
\end{figure}

We use the EK55 dataset to demonstrate the effect of observation time on HOI prediction performance. We present the change of hand trajectory prediction errors with different input sequence lengths $\{2,4,6,8,10\}$, corresponding to the observation time $\{0.5\,\text{s},1.0\,\text{s},1.5\,\text{s},2.0\,\text{s},2.5\,\text{s}\}$. We first use Diff-IP2D trained with $10$ observation frames to implement zero-shot prediction with different sequence lengths. Fig.~\ref{fig:compare_obs}(a) illustrates that the prediction performance drops significantly when the number of observation frames decreases. 
In contrast, once our proposed model is trained from scratch with the predefined observation time, it generates plausible prediction results as Fig.~\ref{fig:compare_obs}(b) shows. Especially when the number of observation frames decreases to $4$, our method still outperforms the baseline which is trained from scratch with $10$ observation frames. This demonstrates the strong generation ability of our diffusion-based approach with limited conditions.

\subsection{Additional Visualization of Object Affordance Prediction on Epic-Kitchens}
\label{sec:add_viz_affordance}
\begin{figure}[h]
  \centering
  \includegraphics[width=1\linewidth]{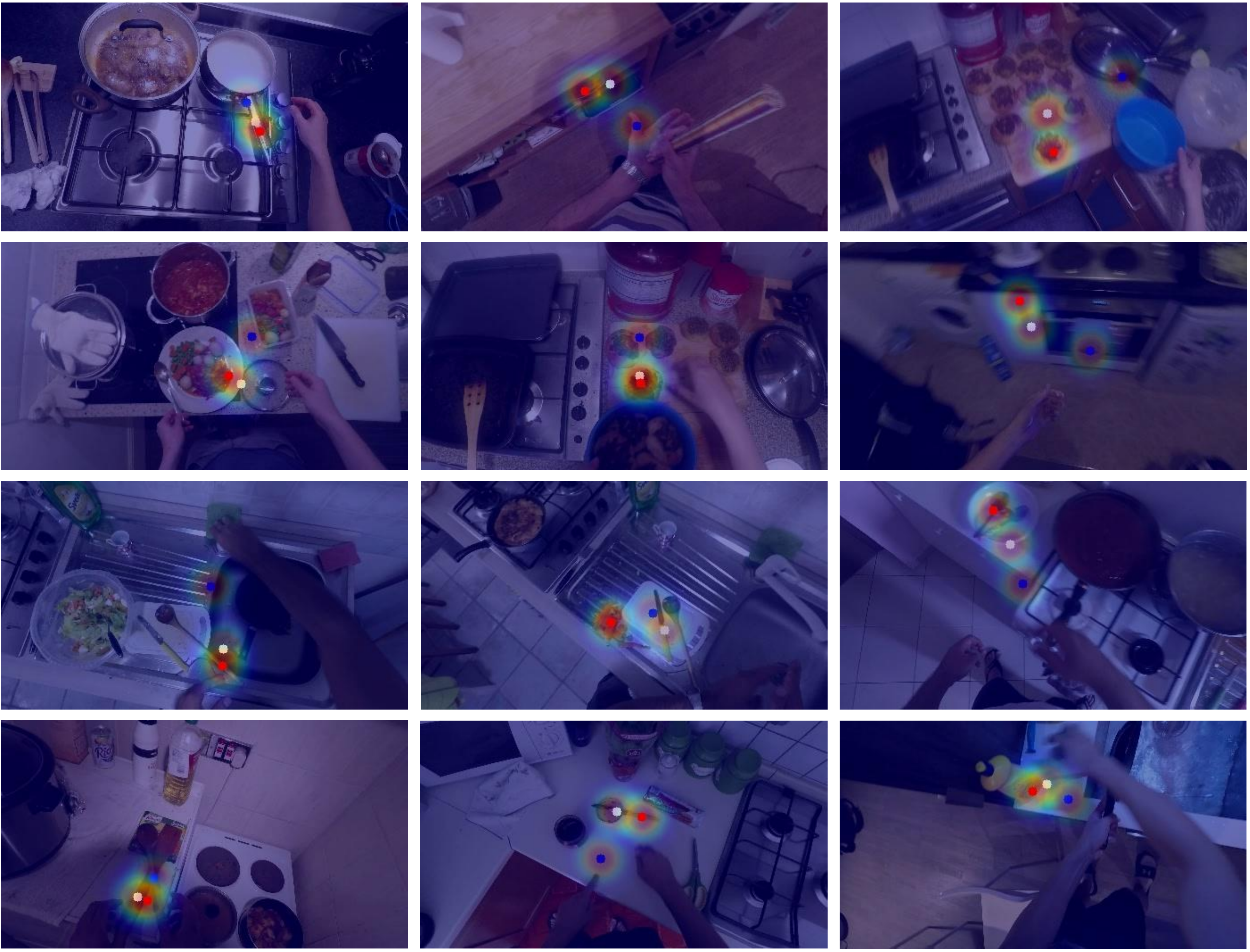}
  \caption{Visualization of object affordance prediction grounded on frames of Epic-Kitchens. The ground-truth contact points are represented by red dots. The contact points predicted by our Diff-IP2D with average distances to the ground-truth points are represented by white dots. The counterparts predicted by OCT \cite{liu2022joint} are represented by blue dots.}
  \label{fig:obj_pred_add}
\end{figure}

We additionally illustrate the predicted contact points with average distances to the ground-truth points on frames of Epic-Kitchens. As Fig.~\ref{fig:obj_pred_add} shows, our proposed method still outperforms the second-best baseline considering the center of 10 predicted candidates.

\begin{figure}[h]
  \centering
 \includegraphics[width=1\linewidth]{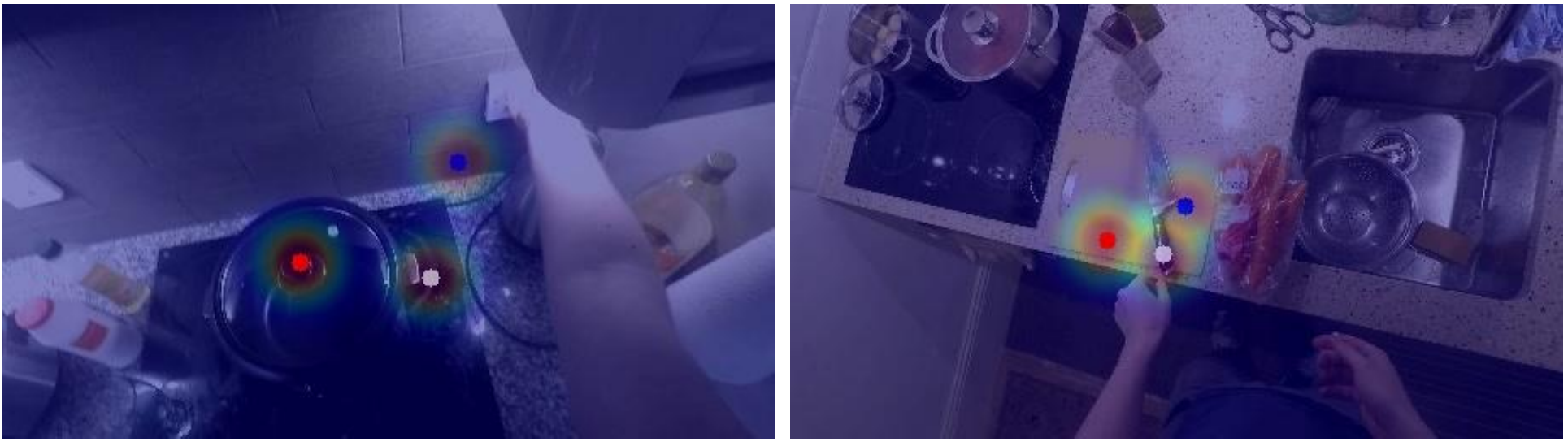}
 \caption{Two additional explanatory cases.}
  \label{fig:failure_cases}
    \vspace{-0.7cm}
\end{figure}
We also provide two cases in which our Diff-IP2D predicts object affordances away from ground truth but more reasonable than the counterparts of the baseline. As Fig.~\ref{fig:failure_cases} shows, our proposed Diff-IP2D focuses more on ``meaningful'' parts of objects such as handles even though its prediction has a similar distance away from ground-truth contact points compared to the baseline.

\begin{table}[t] \small
\setlength{\tabcolsep}{6pt}
\center
\renewcommand\arraystretch{1.1}
\caption{Comparison of performance on hand trajectory prediction on EgoPAT3D-DT}
\begin{tabular}{l|ccc}
\toprule
metric  & OCT~\cite{liu2022joint} & USST~\cite{bao2023uncertainty} & Diff-IP2D$^{\dag}$ (ours)      \\ \cmidrule{1-4} 
ADE~(seen)   & 0.108    & 0.082     &  \textbf{0.076}            \\
FDE~(seen)   & 0.122    & 0.118      & \textbf{0.112}         \\ \midrule
ADE~(unseen)   & 0.091    & 0.060     &  \textbf{0.055}            \\
FDE~(unseen)   & 0.147    & 0.087      & \textbf{0.083}         \\\bottomrule
\end{tabular}\\
\scriptsize Final displacement errors of baselines~\cite{liu2022joint,bao2023uncertainty} are re-evaluated according to the erratum from Bao \textit{et al.}~\cite{bao2023uncertainty} in their open-source repository.
\label{tab:egopat3d}
\vspace{-0.5cm}
\end{table}

\subsection{Evaluation on EgoPAT3D-DT}

We further conduct an additional experiment on a new public dataset EgoPAT3D-DT~\cite{{li2022egocentric,bao2023uncertainty}} to compare the performance of our proposed Diff-IP2D$^{\dag}$ and two state-of-the-art baselines, OCT~\cite{liu2022joint} and USST~\cite{bao2023uncertainty}. There is no affordance annotation in EgoPAT3D-DT and thus we only report the results of hand trajectory prediction. Following the previous work \cite{bao2023uncertainty}, we use the fixed ratio 60\% to split the past and future sequences at 30 FPS. EgoPAT3D-DT encompasses both seen scenes and unseen scenes, where the unseen scenes are only used for testing. We obtain 6356 training sequences, 846 validation sequences, and 1605 test sequences. As can be seen in Tab.~\ref{tab:egopat3d}, our Diff-IP2D$^{\dag}$ conducted on the iter-NAR paradigm with temporal enhancement outperforms the AR baselines on hand trajectory prediction on the EgoPAT3D-DT dataset. The better performance of our proposed approach on the unseen test scenes also demonstrates its solid generalization ability.

\section{Supplementary Technical Details}

\textbf{How the GT future hand trajectories are obtained for training? How good are they?} We follow the GT labels of future hand trajectories from Liu \etal~\cite{liu2022joint}. They use a hand-object detector~\cite{shan2020understanding} to extract hand bounding boxes for each future image. Each bounding box center is projected to the last observation frame (canvas image) using estimated homography. The homography matrix between the future image and the canvas image is obtained by multiplying sequential homography matrices. The projected hand locations in the canvas image plane constitute future hand waypoints for training and testing our model. The GT annotations are high quality because: (1) the hand-object detector achieves around 90\% IOU on egocentric datasets~\cite{shan2020understanding}, (2) low-quality GT hand trajectories are manually removed by Liu \etal~\cite{liu2022joint}, and (3) we have rechecked the quality of GT hand trajectories in this work.

\textbf{How does the model handle the situation when only one hand is visible in the frames?} The above-mentioned hand-object detector identifies the visibility of each hand, providing a side-aware valid mask for our work. During training and testing, we pad zero values to the output features of SOFM at the timestamps when the hand is invisible according to the valid mask. Besides, the mask is also used by MADT while computing self- and cross-attention. Moreover, Hermite spline interpolation is used to fill the missing GT waypoints caused by invisible hands. If one hand, e.g., the left hand, is absent throughout the entire video clip, Diff-IP2D focuses diffusion denoising on the visible side for higher efficiency, as the latent features from our SOFM are side-aware. In these cases, the invisible side cannot be used for both supervision and error calculation.

\textbf{How are the ground truth 10 contact points obtained? How good are they?} In the training process, we use GT contact points from Liu \etal~\cite{liu2022joint}, who exploit skin segmentation and fingertip detection to determine fingertip locations within hand-object bounding boxes. Each training video clip has one GT future contact point per valid side, averaged from detected fingertip locations. Diff-IP2D outputs one possible contact point per valid side after each forward process. For testing, we also follow Liu \etal's \cite{liu2022joint} evaluation pipeline and use their high-quality GT object hotspot annotations from Amazon Mechanical Turk and rechecked by us. Each video clip has 1-5 GT points as the ``contact center'' annotated by workers in the canvas frame to generate GT hotspots, manually avoiding the occlusion problem. To generate object affordance predictions, we perform 10 inference samples per valid side and select the predicted contact point closest to the predicted hand trajectories. These 10 selected points represent sampled estimates of possible next-active object locations, which are used to calculate the object hotspot as affordance.

\end{document}